\documentclass{article}

\PassOptionsToPackage{numbers}{natbib}

\usepackage[main, final]{neurips_2026}

\usepackage[utf8]{inputenc} 
\usepackage[T1]{fontenc}    
\usepackage{hyperref}       
\usepackage{url}            
\usepackage{booktabs}       
\usepackage{amsfonts}       
\usepackage{nicefrac}       
\usepackage{microtype}      
\usepackage{xcolor}         

\definecolor{mydarkblue}{rgb}{0,0.08,0.45}
\hypersetup{
	colorlinks,
	linkcolor={mydarkblue},
	citecolor={mydarkblue},
	urlcolor={mydarkblue}
}

\usepackage{subcaption}
\usepackage{graphicx}
\usepackage{amsmath}
\usepackage{amssymb}
\usepackage{mathtools}
\usepackage{amsthm}
\usepackage[capitalize,noabbrev]{cleveref}
\usepackage[bb=dsserif]{mathalpha}

\theoremstyle{plain}

\theoremstyle{definition}

\newtheorem{example}{Example}[section]
\theoremstyle{remark}

\usepackage{tikz}
\usetikzlibrary{automata,arrows.meta,calc,positioning,decorations.pathreplacing,backgrounds,matrix,arrows,patterns}
\usepackage{bm}
\usepackage{multirow}
\usepackage{multicol}
\usepackage{algorithm}
\usepackage{algorithmic}
\usepackage{enumitem}
\usepackage{wrapfig}

\newcommand*{\given}{\,|\,}

\newcommand{\always}{\mathsf{G}\,}
\newcommand{\event}{\mathsf{F}\,}

\newcommand{\nex}{\mathsf{X}\,}
\newcommand{\until}{\;\mathsf{U}\;}

\newcommand{\gror}{\;|\;}

\DeclareMathOperator*{\argmax}{arg\,max}

\DeclareMathOperator*{\softmax}{softmax}
\DeclareMathOperator*{\E}{\mathbb E}

\DeclarePairedDelimiter\abs{\lvert}{\rvert}%
\DeclarePairedDelimiter\norm{\lVert}{\rVert}%
\makeatletter
\let\oldabs\abs
\def\abs{\@ifstar{\oldabs}{\oldabs*}}
\let\oldnorm\norm
\def\norm{\@ifstar{\oldnorm}{\oldnorm*}}
\makeatother

\usepackage{pifont}
\title{Zero-Shot Instruction Following in RL via\\Structured LTL Representations}

\author{%
Mathias Jackermeier$^{1}$ \quad Mattia Giuri$^{1,2}$ \quad Jacques Cloete$^3$ \quad Alessandro Abate$^1$\\
$^1$Department of Computer Science, University of Oxford\\
$^2$Qube Research and Technologies\\
$^3$Oxford Robotics Institute, University of Oxford
}

\begin{document}

\maketitle

\begin{abstract}
  We study instruction following in multi-task reinforcement learning, where an agent must zero-shot execute novel tasks not seen during training.
  In this setting, linear temporal logic (LTL) has been adopted as a powerful framework for specifying structured, temporally extended tasks.
  While existing approaches successfully train generalist policies, they often struggle to effectively capture the rich logical and temporal structure inherent in LTL specifications.
  In this work, we address these concerns with a novel approach to learn \textit{structured} task representations that facilitate training and generalisation.
  Our method conditions the policy on sequences of Boolean formulae constructed from a finite automaton of the task.
  We propose a hierarchical neural architecture to encode the logical structure of these formulae, and introduce an attention mechanism that enables the policy to reason about future subgoals.
  Experiments in a variety of complex domains demonstrate the strong generalisation capabilities and superior performance of our approach.
\end{abstract}

\section{Introduction}
Recent years have seen remarkable progress in training artificial intelligence agents to follow arbitrary instructions~\citep{luketina2019survey,liu2022GoalConditioned,paglieri2025BALROG,klissarov2025MaestroMotif}. A key consideration in this line of work is the \textit{type} of instruction to provide to the agent. While a long line of research has investigated tasks expressed in natural language~\citep{oh2017ZeroShot,goyal2019Using,hill2020Human,carta2023Grounding}, there recently has been increased interest in training agents to follow instructions specified in \textit{formal} language~\citep{jothimurugan2021Compositional,vaezipoor2021LTL2Action,jackermeier2025deepltl}. Compared to natural language, formal specifications offer several advantages: they facilitate the training process by enabling the automatic construction of task-specific reward functions, they explicitly expose task structure via their compositional syntax, and they are grounded in well-defined semantics. This makes formal instruction languages especially appealing in safety-critical settings, where specifying unambiguous tasks is important~\citep{leon2021Systematic}.

One particular formal language that has emerged as an expressive formalism to specify tasks in reinforcement learning (RL) is \textit{linear temporal logic} (LTL)~\cite{pnueli1977temporal}. LTL instructions are defined over a set of \textit{atomic propositions}, corresponding to high-level events that can hold true or false at each state of the environment. These atomic propositions are combined using logical and temporal operators, which allow for the specification of complex, temporally extended behaviour in a compositional manner. Recent work has exploited the connection between LTL and corresponding automata structures (typically variants of B\"uchi automata~\citep{buchi1966Symposium}), which provide an automatic way to monitor task progress, to train generalist policies capable of executing arbitrary LTL instructions~\citep{qiu2023Instructing,jackermeier2025deepltl,guo2025One}.

In this paper, we develop a novel approach to train generalist LTL-conditioned policies.
Our method is designed to explicitly model the rich logical and temporal structure of LTL specifications, in order to yield \textit{structured} task representations that accelerate learning and enable improved generalisation.
We obtain these representations by translating transitions in the automaton into equivalent Boolean formulae, which provide succinct, compositional descriptions of the logical conditions required to make progress towards a given task.
We propose a hierarchical neural architecture to encode these formulae, and introduce an attention mechanism that allows the policy to reason about future subgoals, enabling more effective planning.
Our main contributions are as follows:
\begin{itemize}[itemsep=0em]
  \item we propose representing LTL instructions as sequences of minimal Boolean formulae constructed from transitions in the corresponding B\"uchi automaton;
  \item we develop a structured representation learning approach to embed these sequences of formulae, based on a hierarchical neural architecture coupled with a temporal attention mechanism;
  \item we introduce a novel environment with complex logical structure, which allows us to systematically study the effectiveness of our approach across challenging LTL specifications;
  \item lastly, we conduct an extensive empirical evaluation demonstrating that our method achieves state-of-the-art results and significantly outperforms existing approaches across a wide range of tasks.
\end{itemize}

\section{Related Work}
Significant research effort has explored LTL as a task specification language for RL agents, with many works training agents to satisfy a \textit{single}, fixed specification~\citep{sadigh2014learning,hasanbeig2018LogicallyConstrained,camacho2019LTL,hahn2019OmegaRegular,bozkurt2020Control,cai2021Reinforcement,shao2023Sample,voloshin2023Eventual,le2024Reinforcement,shah2025LTLConstrained}. In contrast, we aim to learn a general multi-task policy for zero-shot generalisation to unseen tasks.

Several approaches to this problem decompose LTL specifications into subtasks completed sequentially by a goal-conditioned policy~\citep{araki2021Logical,leon2021Systematic,leon2022Nutshell,qiu2023Instructing,liu2024Skill,guo2025One}. While this simplifies the learning process, such methods can exhibit \textit{myopic} behaviour, since attending to a single subgoal at a time may not yield a globally optimal solution. \citet{kuo2020Encoding} instead compose RNNs mirroring the formula structure, though this requires learning a non-stationary policy, which is generally challenging~\citep{vaezipoor2021LTL2Action}.

\citet{vaezipoor2021LTL2Action} introduce LTL2Action, encoding the formula's syntax tree with a GNN and applying LTL progression~\citep{bacchus2000Using} to continuously update the task representation. The primary drawback is that the policy must \textit{learn} the semantics of temporal operators, and infinite-horizon tasks are not supported. Instead, we construct B\"uchi automata to explicitly capture the temporal structure of tasks.

Automata-based representations have previously been explored, but our work differs in how the task is represented and processed.
Instead of learning a policy conditioned myopically on single atomic propositions~\citep{qiu2023Instructing}, or on sequences of unstructured sets of assignments~\citep{jackermeier2025deepltl}, we construct semantically meaningful Boolean formulae from automaton transitions, yielding a more explicit and structured representation of task dynamics. We furthermore incorporate information about future subgoals via a novel attention mechanism.
Finally, recent work~\citep{cloete2026platoltl} explores scaling LTL-guided RL to parameterised predicates, which is orthogonal to our focus on structured task representations.

\section{Background}
\paragraph{Reinforcement Learning.}
We formally model reinforcement learning (RL) environments as Markov decision processes (MDPs). An MDP is defined as a tuple $\mathcal{M} = (\mathcal{S}, \mathcal{A}, p, r, \gamma, \rho_0)$, where $\mathcal{S}$ is the state space, $\mathcal{A}$ is the action space, $p: \mathcal{S} \times \mathcal{A} \times \mathcal{S} \rightarrow [0,1]$ is the transition probability function, $r: \mathcal{S} \times \mathcal{A} \times \mathcal{S} \rightarrow \mathbb{R}$ is the reward function, $\gamma \in [0,1)$ is the discount factor, and $\rho_0$ is the initial state distribution. A (memoryless) policy $\pi: \mathcal{S} \rightarrow \Delta(\mathcal{A})$ maps states to probability distributions over actions. The goal of RL is to find a policy $\pi$ that maximises the expected discounted return $J(\pi) = \mathbb{E}_{\tau \sim \pi} [\sum_{t=0}^{\infty} \gamma^t r_t]$, where $\tau = (s_0, a_0, r_0, s_1, \dots)$ is a trajectory generated by following $\pi$ starting from $s_0 \sim \rho_0$, that is, $a_t\sim\pi(\cdot\given s_t)$, $s_{t+1}\sim p(\cdot\given s_t,a_t)$, and $r_t = r(s_t,a_t,s_{t+1})$. The value function of $\pi$ is defined as $V^\pi(s) = \mathbb{E}_{\tau \sim \pi} [\sum_{t=0}^{\infty} \gamma^t r_t\given s_0 = s]$, i.e.\ the expected discounted return of $\pi$ starting from state $s$.

\paragraph{Linear Temporal Logic.}
We use linear temporal logic (LTL)~\citep{pnueli1977temporal} to specify tasks for RL agents.
LTL formulae are defined over a set of atomic propositions ($AP$), which represent basic properties of the environment such as ``object $A$ is at location $X$'' or ``the agent is in the green region''.
LTL formulae are defined recursively as
$$ \varphi ::= \top \gror \mathsf p \gror \neg \varphi \gror \varphi \land \psi \gror \nex \varphi \gror \varphi \until \psi $$
where $\top$ denotes true, $\mathsf p \in AP$ is an atomic proposition, $\varphi$ and $\psi$ are themselves LTL formulae, $\neg$~(negation) and $\land$ (conjunction) are standard Boolean connectives (from which others like $\lor$, $\rightarrow$, $\leftrightarrow$ can be derived), and $\mathsf X$ (next) and $\mathsf U$ (until) are temporal operators. From these, the common temporal operators $\event\varphi \equiv \top \until \varphi$ (eventually or finally) and $\always \varphi \equiv \neg \event \neg \varphi$ (always or globally) can be derived. Intuitively, $\varphi\until\psi$ is true if $\psi$ becomes satisfied at some time step $t$ and $\varphi$ is true at all previous time steps. Accordingly, $\event \varphi$ simply states that $\varphi$ has to be true \textit{eventually}, whereas $\always \varphi$ requires $\varphi$ to be true at every time step.

Formally, LTL semantics are defined over infinite traces of \textit{truth assignments} $\sigma = \sigma_0 \sigma_1 \sigma_2 \dots$, where $\sigma_t \subseteq AP$ is the set of atomic propositions true at time $t$. We assume a (known) labelling function $L\colon \mathcal{S} \rightarrow 2^{AP}$, which maps MDP environment states $s \in \mathcal{S}$ to corresponding sets of true atomic propositions. A trajectory $\tau = (s_0, a_0, s_1, \dots)$ satisfies an LTL formula $\varphi$, denoted $\tau \models \varphi$, if its trace $L(s_0)L(s_1)\dots$ satisfies $\varphi$ according to LTL semantics (see \cref{app:ltl_semantics}). For example, the formula $\neg\mathsf a\until \mathsf b$ is satisfied by exactly the trajectories where eventually $\mathsf b$ is true at some time step $t$ (i.e.\ $\mathsf b\in L(s_t)$), and $\mathsf a$ is false at all time steps before. LTL provides a formal and structured way to define complex tasks to RL agents, such as ``eventually reach region A, and if you pick up an object on the way, you must also visit region B while avoiding region C''.
The problem of training a generalist LTL-guided policy can be formalised by defining a distribution $\mathcal D$ over LTL tasks.
We then aim to compute the optimal task-conditioned policy
\begin{equation}\label{eq:objective}
  \pi^* \in \argmax_\pi \E_{\substack{\varphi \sim \mathcal D, \\ \tau \sim \pi \given \varphi}}\big[ \mathbb{1}[ \tau \models \varphi ] \big].
\end{equation}

\paragraph{Büchi Automata.}
The semantics of LTL can equivalently be captured by Büchi automata~\citep{buchi1966Symposium}, which provide a more explicit way to reason about tasks. We here focus on \textit{limit-deterministic} B\"uchi automata (LDBAs)~\citep{sickert2016LimitDeterministic}, which are defined as tuples $\mathcal B = (\mathcal Q, q_0, \Sigma, \delta, \mathcal E, \mathcal F)$.
\begin{wrapfigure}{r}{.44\textwidth}
  \centering
  \resizebox{.35\textwidth}{!}{
    \begin{tikzpicture}[->,>=stealth',shorten >=1pt,auto,semithick,]
      \node[state,initial] (q0) {$q_0$};
      \node[state] (q1) [above right=0.4cm and 1.4cm of q0] {$q_1$};
      \node[state,accepting] (qx) [right=1.2cm of q1] {$q_2$};
      \node[state,accepting] (q2) [below right=0.4cm and 1.4cm of q0] {$q_3$};
      \node[state] (q3) [right=1.2cm of q2] {$\bot$};
      \path (q0) edge [loop above] node {$\top$} (q0)
      (q0) edge node {$\varepsilon_1$} (q1)
      (q1) edge node {$\{\mathsf{red}\}$} (qx)
      (q1) edge [loop above] node {$\{\mathsf{green}\},\emptyset$} (q1)
      (qx) edge [loop above] node {$\top$} (qx)
      (q0) edge [below,pos=0.3] node {$\varepsilon_2$} (q2)
      (q2) edge [loop above] node {$\{\mathsf{green}\}$} (q2)
      (q2) edge [above] node {$\{\mathsf{red}\},\emptyset$} (q3)
      (q3) edge [loop above] node {$\top$} (q3);
    \end{tikzpicture}
  }
  \caption{LDBA for the LTL formula $\event \mathsf{red} \lor (\event\always \mathsf{green})$.}
  \label{fig:ldba}
\end{wrapfigure}
$\mathcal Q = \mathcal Q_I \uplus \mathcal Q_A$ is a finite set of states partitioned into two subsets, $q_0\in\mathcal Q$ is the initial state, $\Sigma = 2^{AP}$ is a finite alphabet, $\delta\colon \mathcal Q\times(\Sigma\cup\mathcal E)\to \mathcal Q$ is the transition function, and $\mathcal F$ is the set of accepting states. We require that $\mathcal F\subseteq \mathcal Q_A$ and $\delta(q,\cdot)\in\mathcal Q_A$ for all $q\in\mathcal Q_A$. The only way to transition from $\mathcal Q_I$ to $\mathcal Q_A$ is by taking an $\varepsilon$-transition $\varepsilon\in\mathcal E$ (a.k.a.\ \textit{jump} transition), which does not consume any input. Given an input trace $\sigma$, a \textit{run} of $\mathcal B$ is an infinite sequence of states in $\mathcal Q$ starting in $q_0$ and respecting the transition function $\delta$. A trace is \textit{accepted} by $\mathcal B$ if there exists a run that infinitely often visits accepting states. There exist standard algorithms, e.g.\ \citep{sickert2016LimitDeterministic}, to construct an LDBA $\mathcal B_\varphi$ from an LTL formula $\varphi$ such that $\mathcal B_\varphi$ accepts exactly the traces satisfying $\varphi$.

\begin{example}
  \cref{fig:ldba} shows an LDBA for the formula $\event \mathsf{red} \lor (\event\always \mathsf{green})$. In $q_0$, the agent can either choose $\varepsilon_1$ to transition to $q_1$, from which it reaches the accepting state $q_2$ upon visiting an MDP state $s$ with $\mathsf{red}\in L(s)$. By following $\varepsilon_2$, it instead has to stay in a $\mathsf{green}$ state forever. Note that we assume here that $\mathsf{red}$ and $\mathsf{green}$ cannot be true together.
\end{example}

\begin{figure*}[t]
  \centering
  \includegraphics[scale=0.7]{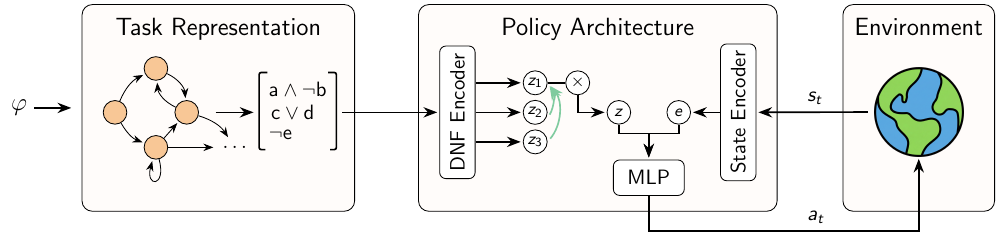}
  \caption{Overview of our method. \textit{(Left)} Given an LTL instruction $\varphi$, we construct an LDBA and extract sequences of Boolean formulae representing the current automaton state. \textit{(Right)} The policy is conditioned on a single formula sequence. It processes formulae with a hierarchical DNF encoder and a temporal attention mechanism to reason about future subgoals. The latent task representation $z$ is concatenated with the encoded MDP state $s_t$ and mapped to an action $a_t$ via an actor MLP.}
  \label{fig:method-overview}
\end{figure*}

B\"uchi automata are well-suited to represent LTL instructions, since they explicitly capture the memory required to execute a given task via their states. It is hence common to consider policies $\pi\colon \mathcal S\times\mathcal Q\to\Delta(\mathcal A)$ conditioned not only on the current MDP state $s$, but also on the current LDBA state $q$~\citep{hasanbeig2018LogicallyConstrained,hahn2019OmegaRegular}. After each environment interaction, the automaton state is updated according to the transition function $\delta$ and the currently true propositions $L(s)$. Over the course of a trajectory, the policy can adjust its behaviour as the LDBA state changes. For example, in \cref{fig:ldba} the policy would aim to make $\mathsf{red}$ true in $q_1$, but in $q_3$ instead try to stay in a $\mathsf{green}$ state.

\section{Structured LTL Representations}
In single-task RL, it is straightforward to condition the policy on LDBA states. Since the automaton remains fixed throughout training and evaluation, one can simply train separate state-specific sub-policies, and choose the appropriate one at test time (e.g.\ \citep{jothimurugan2021Compositional}).
However, this approach is unfeasible in multi-task RL, since the policy must generalise to novel tasks (and therefore automata states) that it has never encountered during training. The key challenge is thus to find a suitable representation of LDBA states that (i) enables such generalisation, and (ii) is amenable to policy learning.

Recent work has explored various representations, such as conditioning on single atomic propositions~\citep{qiu2023Instructing}, sequences of unstructured sets of assignments~\citep{jackermeier2025deepltl}, or directly encoding the LTL formula syntax tree~\citep{vaezipoor2021LTL2Action}.
However, these approaches struggle to effectively capture the rich logical and temporal structure of tasks, particularly in environments where many propositions can hold simultaneously.

As illustrated in \cref{fig:method-overview}, we instead propose representing LDBA states as \textit{sequences of Boolean formulae}.
This representation explicitly exposes the logical structure of tasks, enabling the policy to generalise compositionally.
For example, it can leverage knowledge about achieving propositions $\mathsf{a}$ and $\mathsf{b}$ individually to also achieve $\mathsf{a} \land \mathsf{b}$.
Furthermore, since the policy is conditioned on a \textit{sequence} of such formulae, it can non-myopically reason about future subgoals, which may affect how it approaches the current step.

Intuitively, these sequences of Boolean formulae align with \textit{accepting runs} in the LDBA, informing the policy of the high-level steps required to complete the task.
We design a procedure to automatically construct such sequences from the LDBA (\cref{sec:boolean-formulae}), and propose a specialised neural network architecture that leverages the rich information in these sequences to yield \textit{structured} learned task representations (\cref{sec:policy}).
Further details on the training procedure are given in \cref{sec:training}.
Finally, \cref{sec:test-time} describes the test-time execution, where the policy is conditioned on sequences extracted from the current task's LDBA and updates dynamically as the automaton state evolves.


\subsection{Representing LDBA States with Boolean Formulae}\label{sec:boolean-formulae}
Given an LDBA $\mathcal B_\varphi$ and state $q$, we aim to identify behaviour that will lead to satisfying the LTL instruction.
Since $\varphi$ is satisfied exactly when the agent visits accepting states infinitely often, we first enumerate possible ways to reach an \textit{accepting cycle} in $\mathcal B_\varphi$, i.e.\ a cycle with at least one accepting state.
This can be achieved with a simple depth-first search (DFS) starting from $q$ (Algorithm 1 of \citet{jackermeier2025deepltl}; see \cref{app:alg} for details).

The DFS yields a set of accepting runs $\rho_i = (q, q_1, \dots)$ that correspond to different ways of achieving $\varphi$.
For each run $\rho_i$, we construct a sequence of Boolean formulae that captures the high-level goals the agent must complete in order to follow the run, and hence satisfy the LTL task.
Specifically, in order to transition from $q_i$ to $q_{i+1}$, the agent must achieve an assignment $\alpha_i\subseteq AP$ of atomic propositions that satisfies the transition condition $\delta(q_i, \alpha_i) = q_{i+1}$, while avoiding transitions to other states (excluding self-loops). We represent this with two Boolean formulae $\beta_i^+$ and $\beta_i^-$ that satisfy
\begin{align*}
  \forall \alpha\in\mathbb A.\,\alpha\models\beta_i^+ & \iff \delta(q_i,\alpha) = q_{i+1},                \\
  \forall \alpha\in\mathbb A.\,\alpha\models\beta_i^- & \iff \delta(q_i,\alpha) \not\in \{q_i, q_{i+1}\},
\end{align*}
where $\mathbb A = \{L(s) : s\in\mathcal S\} \subseteq 2^{AP}$ is the set of possible assignments in the MDP (which we assume known). Intuitively, $\beta_i^+$ is only true for the assignments that transition from $q_i$ to $q_{i+1}$, whereas $\beta_i^-$ captures the assignments that must be avoided in order to avoid transitioning to other states.
Boolean formulae can be much more succinct than explicitly enumerating sets of assignments, as we demonstrate with the following example.

\begin{example}
  Consider an MDP with propositions $AP = \{a,b,c,d\}$ in which all combinations of propositions can hold true at the same time (i.e.\ $\mathbb A = 2^{AP}$). Assume we have an LDBA in which we can transition from state $q_0$ to $q_1$ via any of the assignments in the set
  \begin{equation*}
    A = \bigl\{ \{a\}, \{a, b\}, \{a, d\}, \{a, b, d\}\bigr\}.
  \end{equation*}
  The formula $\beta_0^+ \equiv a\land\neg c$ succinctly represents exactly this set of assignments.
\end{example}

\paragraph{Constructing Boolean formulae.}
We construct formulae $\beta_i^+$ and $\beta_i^-$ for each step $i$ in a run as follows:
we first identify the relevant sets of assignments $A_i^+$ and $A_i^-$ from the transition function $\delta$, that is, $A_i^+ = \{\alpha\in\mathbb A : \delta(q_i,\alpha)=q_{i + 1}\}$ and $A_i^- = \{\alpha\in\mathbb A : \delta(q_i,\alpha)\not\in \{q_i, q_{i + 1}\}\}$.
We could then trivially construct formulae in \textit{disjunctive normal form} (DNF) by exhaustively building a formula for each assignment.
However, this representation is as verbose as simply listing the assignments themselves; instead, we aim to find small, semantically meaningful formulae that capture the assignments while enabling generalisation at test time.

Constructing minimal Boolean formulae is a well-studied problem in formal logic and the minimisation of digital circuits.
We employ the widely used Quine-McCluskey algorithm \citep{quine1952problem,mccluskey1956minimization} to compute minimal DNF representations of $\beta_i^+$ and $\beta_i^-$.
This algorithm takes as input a set of satisfying assignments $S$ and a set of non-satisfying assignments $N$ to produce a minimal (in terms of number of disjunctions and conjunctions) DNF formula that evaluates to true for assignments in $S$ and to false for assignments in $N$.
We set $S = A_i^+$ (or $A_i^-$) and $N = \mathbb A\setminus S$ to yield the formulae $\beta_i^+$ and $\beta_i^-$, respectively.

Note that an accepting run may contain an $\varepsilon$-transition representing a non-deterministic choice.
These transitions do not consume input and hence require no Boolean formula; we instead represent them with a special token $\beta_\varepsilon$ that we handle separately in our policy architecture (see \cref{sec:policy}).

\subsection{Structured Representation Learning}
\label{sec:policy}
Effectively conditioning a policy on the sequences derived in \cref{sec:boolean-formulae} presents two challenges: (i) the Boolean formulae possess hierarchical logical structure that unstructured embeddings fail to capture, and (ii) the sequences vary in length depending on the task's complexity. To address these challenges, we propose a hierarchical architecture to encode the structure of individual formulae and a temporal attention mechanism for the sequence of subgoals.
This is illustrated in \cref{fig:method-overview} (right).

Formally, we denote by $\zeta = ((\beta_1^+,\beta_1^-),(\beta_2^+,\beta_2^-),\ldots)$ a sequence of Boolean formula pairs extracted from an accepting run.
Our policy $\pi(a\given s,\zeta)$ takes as input the current MDP state $s$ and such a target sequence $\zeta$.
Note that while there may be multiple accepting runs for a given LDBA state, the policy is conditioned on a \textit{single} formula sequence at any given time; we detail the run selection procedure in \cref{sec:test-time}.

\paragraph{Embedding Boolean formulae.}
There are a number of straightforward choices to embed Boolean formulae, such as token-level sequence encoders (e.g.\ LSTMs or GRUs) or structure-aware encoders, e.g.\ applying a GNN to a syntax tree representation of the formula~\citep{evans2018Can, crouse2020Improving}.
Rather than using such general methods, we exploit two key features of the formulae encountered by our policy:
(i) they only contain propositions from $AP$, and (ii) they are guaranteed to be in DNF by construction.

Specifically, a formula in DNF consists of a \textit{disjunction} of one or more \textit{clauses}, i.e.\ has the form $\bigvee_{i=1}^n\bigwedge_{j=1}^{m_i} l_{ij}$, where $l_{ij}$ is a possibly negated atomic proposition.
We process DNF formulae in a hierarchical fashion as follows.
Our policy associates a trainable embedding $\bm e_a\in\mathbb R^d$ with each proposition $a\in AP$.
Negated propositions are first processed with a learned linear projection layer, and literals in a clause are then aggregated into a single representation via a permutation-invariant set encoder (since the order of literals does not affect the semantics).
In particular, we use a simple DeepSets encoder~\citep{zaheer2017Deep} consisting of a sum followed by a learned non-linear transformation.
The resulting clause embeddings are aggregated again on the disjunction level by another DeepSets encoder, yielding the overall formula representation $\bm z$.
Formally, we have
\begin{equation*}
  \bm z = f_{\lor}\left(\sum_{i=1}^n f_{\land}\left( \sum_{j=1}^{m_i} \Pi(\bm{e}_{a_{ij}})\right) \right),
\end{equation*}
where $f_\lor$ and $f_\land$ are the non-linear transformations, and $\Pi$ denotes the projection layer if $l_{ij}$ is negated, and the identity function otherwise.
Finally, note that this does not apply to the special $\beta_\varepsilon$ token, for which we instead learn a separate embedding. We give more details on how we handle $\varepsilon$-transitions below.

\paragraph{Attending to future subgoals.}
In order to produce optimal behaviour, the policy may need to consider future subgoals rather than myopically focusing only on the first step of the sequence $\zeta$.
To illustrate, consider a simple sequential task where the agent needs to achieve $\mathsf a$ and then $\mathsf b$.
Depending on the underlying MDP, there may be different ways of making $\mathsf a$ true, some of which might render $\mathsf b$ infeasible (e.g.\ if the agent enters a room in order to achieve $\mathsf a$ that locks upon entry).
At the same time, the first step of the sequence generally contains the most important information, since this is the goal the agent is currently trying to achieve.
We model this intuition with a temporal attention mechanism that allows the policy to contextualise the representation of the current step by attending to future subgoals.

In particular, let $\bm z_i^+$ and $\bm z_i^-$ denote the embedding of $\beta_i^+$ and $\beta_i^-$, respectively.
We first concatenate these embeddings to obtain the overall encoding of step $i$ as $\bm z_i = [\bm z_i^+; \bm z_i^-]$.
We use the first step $\bm z_1$ as the query to attend to the entire sequence using single-head scaled dot-product attention~\citep{vaswani2017Attention}, which computes the output
\begin{equation*}
  \bm h = \left[\softmax\left(\frac{\bm qK^\top}{\sqrt{d_k}}\right)V\right]W^O,
\end{equation*}
where $\bm q = \bm z_1 W^Q$ is the query vector, and $K, V\in\mathbb R^{T\times d_k}$ are the key and value matrices obtained by projecting the sequence embeddings $\bm z_{1:T}$ using $W^K$ and $W^V$, respectively. $W^O$ is the output projection matrix.
Intuitively, $\bm h$ captures how future formulae in the sequence modulate the current goal $\bm z_1$.
We obtain the final contextualised goal representation as $\tilde{\bm z}_1 = \text{LayerNorm}(\bm z_1 + \bm h)$~\citep{ba2016Layer}.
Since pure attention does not preserve sequential information, we employ ALiBi~\citep{press2021Train} to bias the attention towards nearby steps.
Specifically, before applying the softmax, we subtract a linear penalty $m\cdot (i-1)$ from the unnormalised attention score for step $i$ (where $i \geq 2$), with $m$ being a fixed hyperparameter. This augments the embeddings with positional information and captures the intuition that steps further in the future tend to have less influence on the immediate goal.

Finally, note that formula sequences constructed from accepting runs are technically infinite.
We hence truncate sequences after repeating the accepting cycle a finite number of times~\citep{jackermeier2025deepltl}.
Crucially, the policy is still always conditioned on a complete formula sequence leading to an accepting cycle.

\paragraph{Full policy architecture.}
As depicted in \cref{fig:method-overview} (right), our policy not only processes the formula sequence, but also encodes the current MDP state via a \textit{state encoder}.
This is either a multilayer perceptron (MLP) or convolutional neural network (CNN), depending on the environment.
The contextualised formula sequence embedding $\tilde{\bm z}_1$ is concatenated with the encoded MDP state and fed into a final \textit{actor} MLP, which produces a distribution over actions.

We handle $\varepsilon$-transitions in the LDBA by augmenting the action space of the policy with designated $\varepsilon$-actions, as commonly done~\citep{hasanbeig2018LogicallyConstrained,voloshin2023Eventual}.
These do not perform an action in the MDP, but instead simply update the current LDBA state to the target of the corresponding $\varepsilon$-transition, hence updating the product MDP state $(s,q)$.
Our policy produces a mixture distribution, where it takes an $\varepsilon$-action with a learned parameter $p$ and otherwise samples from its distribution over MDP actions.

\subsection{Training Procedure}\label{sec:training}
We optimize the parameters of our model end-to-end via goal-conditioned RL~\citep{liu2022GoalConditioned}.
That is, we sample a random Boolean formula sequence $\zeta$ at the beginning of each training episode, which sometimes contain the $\beta_\varepsilon$-token.
We then execute a rollout with the policy conditioned on $\zeta$, assigning a positive reward of $+1$ to an episode if the agent successfully sequentially satisfies all formulae $\beta_i^+$.
If the agent at any step $i$ instead satisfies the currently active $\beta_i^-$, we assign a negative reward of $-1$.
This reward structure encourages the policy to satisfy formula sequences that lead to accepting cycles, and hence approximates \cref{eq:objective}.
We use \textit{proximal policy optimization} (PPO)~\citep{schulman2017Proximal} to jointly optimise the policy and learn a value function $V^\pi$.
To further improve training convergence in practice, we follow \citet{jackermeier2025deepltl} and use \textit{curriculum learning} to gradually increase the complexity of formula sequences encountered during training (see \cref{app:curriculum} for details).


\subsection{Test-time Policy Execution}\label{sec:test-time}
At test time, we are given a new LTL instruction $\varphi$ and build a corresponding LDBA $\mathcal B_\varphi$.
We then perform the DFS described in \cref{sec:boolean-formulae} to extract a set of Boolean formula sequences $\{\zeta_1, \ldots, \zeta_n\}$ from the initial state $q_0$.
We use the trained \textit{value function} $V^\pi$ to select the formula sequence that the policy is approximately most likely able to achieve from the current MDP state $s$, i.e.\ we compute
$
  \zeta^* = \argmax_{\zeta_i} V^\pi(s,\zeta).
$
Intuitively, $\zeta^*$ corresponds to the path through the automaton with the highest likelihood of success. The value function, which is a by-product of the training algorithm, naturally estimates (a lower bound of) this likelihood and is hence an appropriate choice for selecting a formula sequence.
We then execute the trained policy $\pi$ conditioned on $\zeta^*$, i.e.\ continuously sample actions $a_t\sim\pi(\cdot\given s_t,\zeta^*)$, until the LDBA state changes, upon which we recompute and update the formula sequence used to condition the policy.

\begin{figure*}[t]
  \centering
  \captionsetup[subfigure]{justification=centering}
  \begin{subfigure}[b]{0.24\textwidth}
    \centering
    \includegraphics[width=\textwidth]{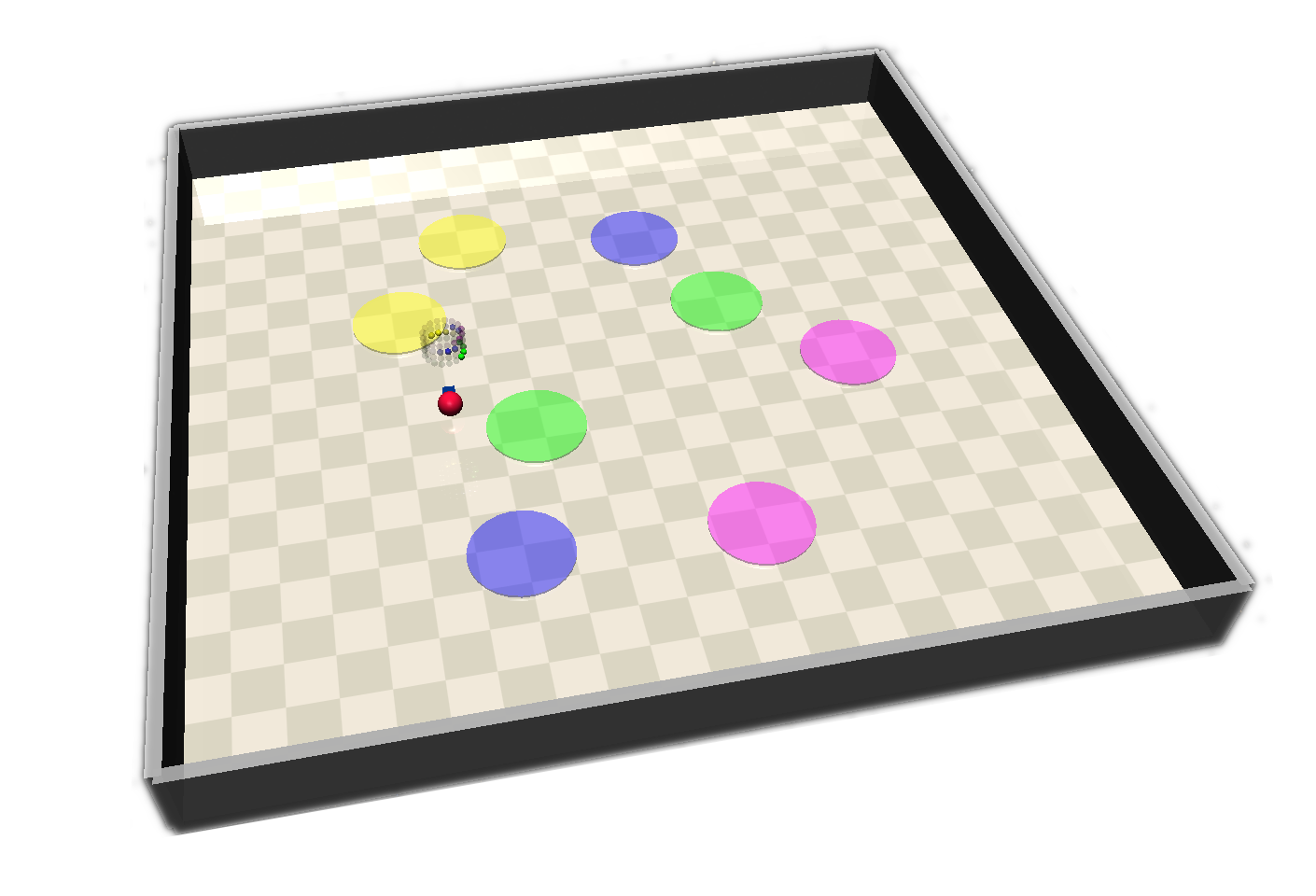}
    \caption{ZoneEnv-NM}
    \label{fig:zones}
  \end{subfigure}
  \hfill
  \begin{subfigure}[b]{0.24\textwidth}
    \centering
    \includegraphics[width=.8\linewidth]{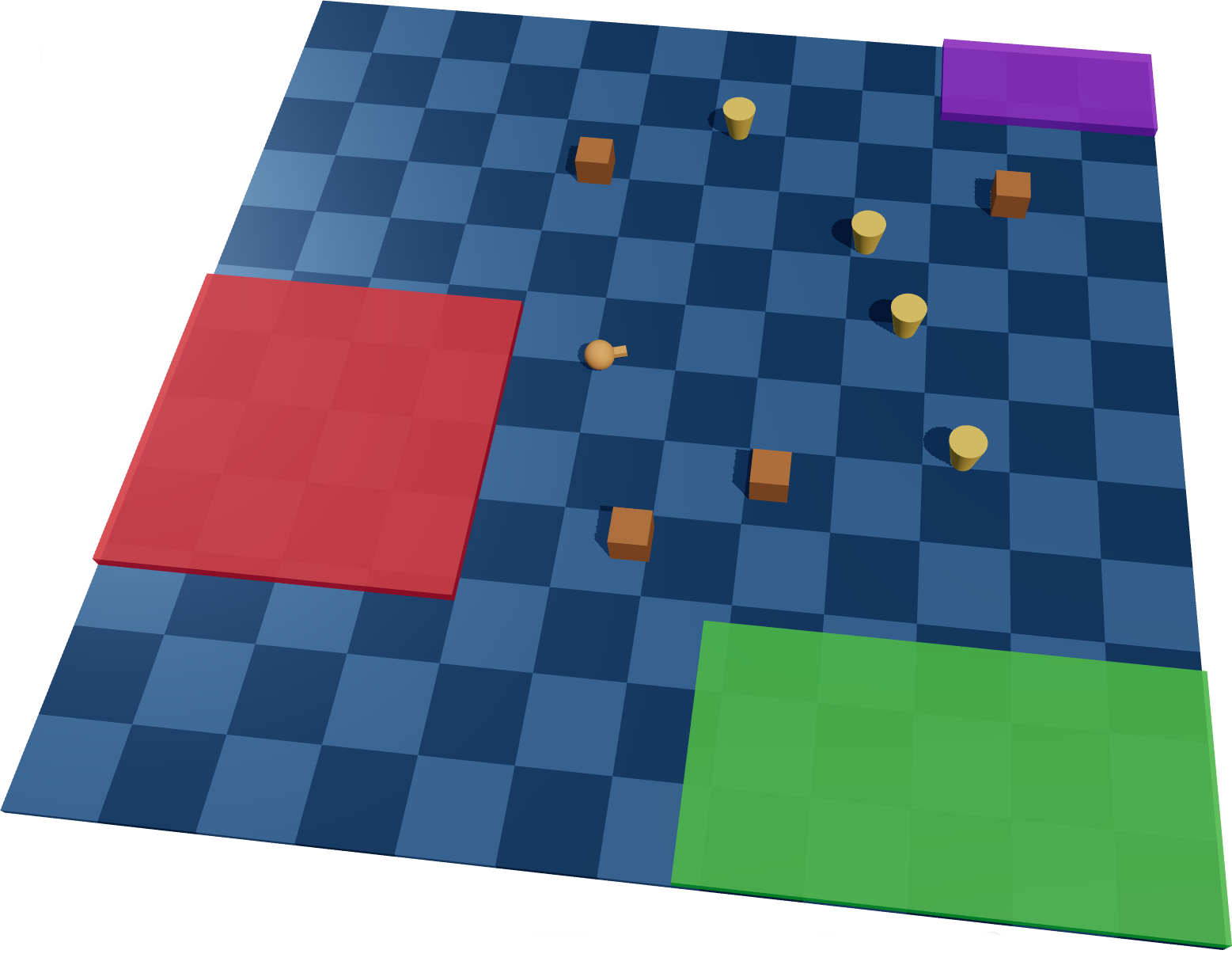}
    \caption{WarehouseEnv}
    \label{fig:warehouse-env}
  \end{subfigure}
  \hfill
  \begin{subfigure}[b]{0.24\textwidth}
    \centering
    \includegraphics[width=.7\linewidth,trim={54bp 39bp 1104bp 39bp},clip]{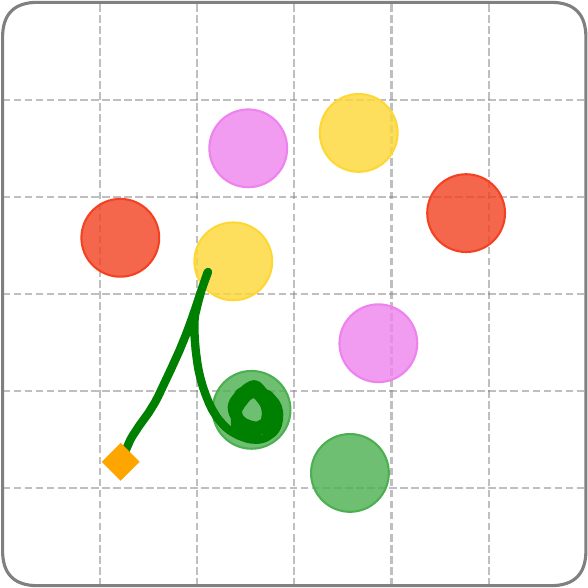}
    \caption{$\event\mathsf{yellow}\wedge\event\always\mathsf{green}$}
    \label{fig:zones-example-traj}
  \end{subfigure}
  \hfill
  \begin{subfigure}[b]{0.24\textwidth}
    \centering
    \includegraphics[width=.705\linewidth]{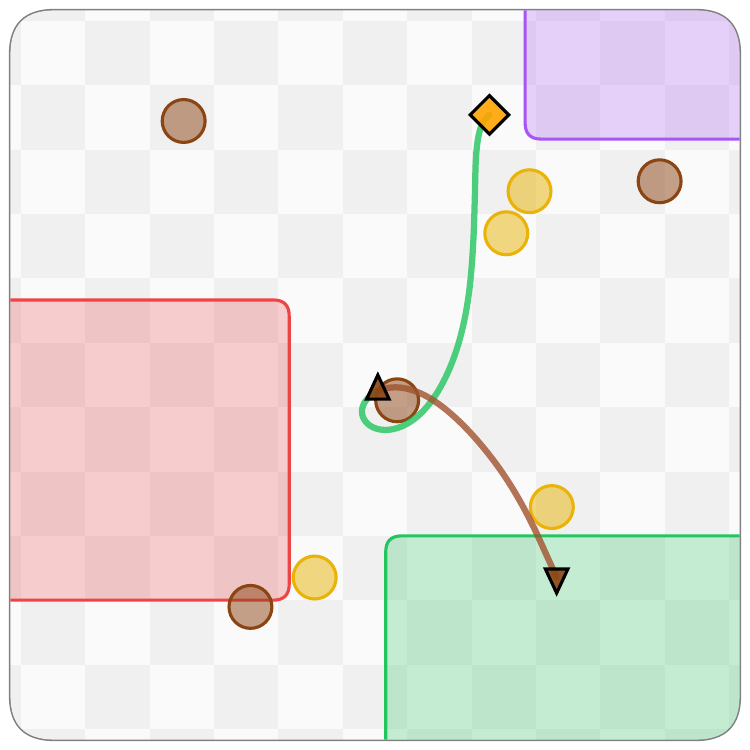}
    \caption{Deliver crate to B (\ref{eq:formula})}
    \label{fig:warehouse-example-traj}
  \end{subfigure}

  \caption{Environment visualisations (left) and example StructLTL trajectories (right). In ZoneEnv-NM, the agent navigates between coloured zones. In Warehouse, the agent moves crates and vases between different regions. Both environments provide high-dimensional lidar observations and continuous movement actions. The orange diamond indicates the starting position of the agent, and triangles pointing upwards and downwards indicate picking up and dropping an object, respectively.}
  \label{fig:combined-envs}
\end{figure*}

\section{Experiments}
\label{sec:experiments}
We evaluate our method across diverse environments, including high-dimensional domains with continuous action spaces.
We aim to answer the following research questions:\footnote{Our code is available at \href{https://github.com/mathiasj33/structltl}{github.com/mathiasj33/structltl}.}
\textbf{(Q1)} Can our approach effectively zero-shot generalise to unseen LTL instructions, particularly those requiring compositional reasoning?
\textbf{(Q2)} Can our method leverage global task information for non-myopic execution?
\textbf{(Q3)} How does our method compare to state-of-the-art baselines in terms of sample efficiency and asymptotic performance?


\paragraph{Environments.}
We conduct experiments in \textit{ZoneEnv} \citep{vaezipoor2021LTL2Action} (\cref{fig:zones}), a well-established high-dimensional robotic navigation environment, in which the agent observes differently coloured zones via a lidar sensor.
The colours correspond to the atomic propositions, which hold true when the agent is located in a corresponding zone.
To increase the complexity of the environment, we introduce state-modifying behaviour requiring non-myopic reasoning (denoted as \textit{ZoneEnv-NM}):
when the agent first enters a purple zone, all green zones disappear for the rest of the episode.
The agent must learn to reason about this behaviour in order to successfully complete complex tasks.

We furthermore introduce the \textit{WarehouseEnv} environment (\cref{fig:warehouse-env}), which admits complex logical specifications.
It consists of an agent navigating fixed regions while interacting with \textit{crates} and \textit{vases} that are randomly scattered across the environment.
The environment features a hybrid action space: the agent can both apply a continuous movement action and choose from a set of discrete actions to pick up and drop objects.
Notably, WarehouseEnv supports complex combinations of atomic propositions holding true at the same time.
For example, if the agent is in region~A while carrying both a vase and crate, all three of these propositions hold true.
Completing tasks in the warehouse environment thus requires the agent to reason compositionally about interactions between propositions.
See \cref{app:environments} for further details on the environments.

\paragraph{Baselines.}
We compare our approach, named StructLTL, to state-of-the-art approaches for learning generalist LTL-conditioned policies.
LTL2Action \citep{vaezipoor2021LTL2Action} directly encodes the syntax tree of a given specification using a GNN\@.
We also compare to DeepLTL \citep{jackermeier2025deepltl}, the state-of-the-art non-myopic approach based on B\"uchi automata.
Finally, we include the state-of-the-art myopic method GenZ-LTL~\citep{guo2025One}, which simplifies the learning problem with hand-designed observation reduction and fusion mechanisms.
Since these are not applicable in the warehouse environment (see \cref{app:hyperparameters} for further discussion), we also compare to a one-hot subgoal variant of GenZ-LTL, denoted as GenZ-LTL$^\dagger$.
Hyperparameters and further details are given in \cref{app:hyperparameters}.

\paragraph{Tasks.}
We consider diverse tasks of varying complexity to evaluate the trained policies.
We differentiate between \textit{finite-horizon} (i.e.\ co-safety) tasks, which can be completed in a finite number of steps, and \textit{infinite-horizon} tasks, which specify recurrent behavior that the agent must execute indefinitely.
Our tasks span a wide range of environment-specific behaviour, such as reaching and avoiding different zones in ZoneEnv-NM, or transporting objects between regions in the Warehouse environment.
They furthermore make use of complex LTL constructs, such as implications to specify \textit{response} tasks and nested temporal operators.
As an example, task $\varphi_{15}$ of our evaluation is given by
\begin{equation}\label{eq:formula}
  (\mathsf{crate} \rightarrow (\mathsf{crate} \until \mathsf{region\_b}))
  \until (\mathsf{crate} \land \mathsf{region\_b} \land (\mathsf{region\_b} \until (\neg \mathsf{crate} \land \mathsf{region\_b}))),
\end{equation}
which can be read as ``Pick up a crate and place it in region B\@. After you have picked up the crate, you must not put it down again, until you have reached the region.''
We provide a full list of evaluation specifications in \cref{app:tasks}.

\begin{figure*}
  \centering
  \includegraphics[width=\textwidth]{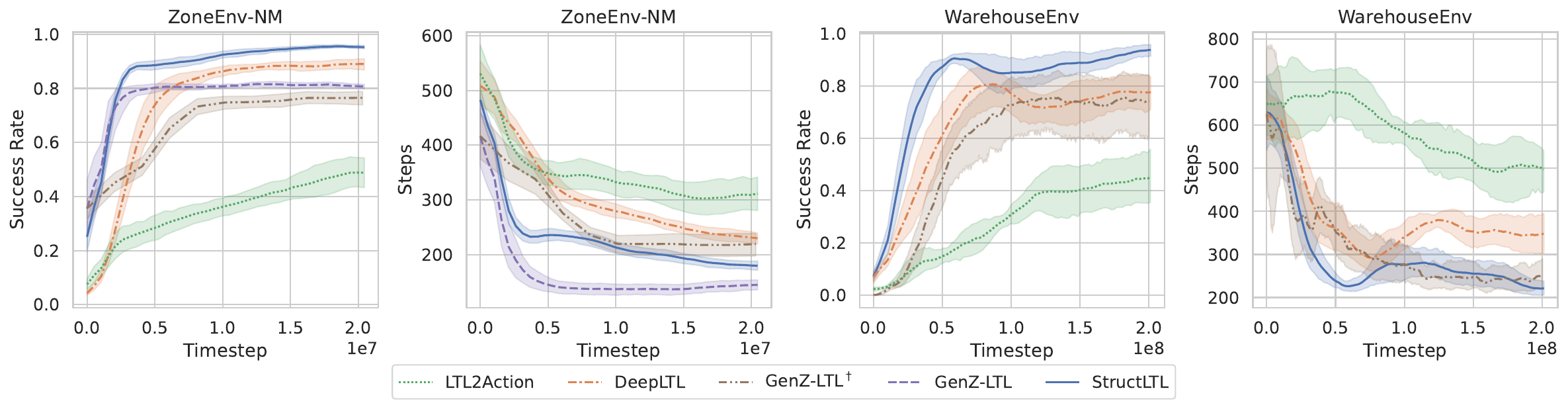}
  \caption{Evaluation curves on finite-horizon tasks over training. We report averages over the entire set of finite-horizon evaluation tasks, computed over 50 episodes per task. Shaded areas indicate 95\% confidence intervals over 10 random seeds.}
  \label{fig:curves}
\end{figure*}

\begin{table*}[t]
    \caption{Evaluation results of trained policies on finite- and infinite-horizon tasks. Results are averaged over 10 seeds and 512 episodes per seed, with 95\% bootstrapped confidence intervals. \textemdash{} indicates not applicable. See \cref{app:tasks} for details on the task specifications.}
    \label{tab:main-results}
    \begin{center}
        \begin{small}

            \newcommand{\metric}[3]{%
                $#1_{\scriptscriptstyle #2}^{\scriptscriptstyle #3}$%
            }

            \resizebox{\textwidth}{!}{%
                \begin{tabular}{ll rrrrr l rrrr}
                    \toprule
                     & \multicolumn{6}{c}{\textbf{Finite Horizon}} & \multicolumn{5}{c}{\textbf{Infinite Horizon}} \\
                    \cmidrule(lr){2-7} \cmidrule(l){8-12}

                     & & \multicolumn{5}{c}{{Success Rate} ($\uparrow$)} & & \multicolumn{4}{c}{{Average Visits ($\uparrow$)}} \\
                    \cmidrule(lr){3-7} \cmidrule(l){9-12}

                     & $\varphi$ & LTL2Action & DeepLTL & GenZ-LTL$^\dagger$ & GenZ-LTL & StructLTL & $\psi$ & DeepLTL & GenZ-LTL$^\dagger$ & GenZ-LTL & StructLTL \\
                    \midrule
                    \multirow{8}{*}{\rotatebox[origin=c]{90}{ZoneEnv-NM}}

                     & $\varphi_{1}$
                     & \metric{0.18}{-0.07}{+0.10} & \metric{0.87}{-0.03}{+0.03} & \metric{0.81}{-0.04}{+0.04} & \metric{0.92}{-0.01}{+0.01} & \metric{\textbf{0.93}}{-0.01}{+0.01}
                     & $\psi_{1}$ & \metric{2.16}{-00.36}{+00.37} & \metric{1.96}{-0.47}{+0.47} & \metric{\textbf{4.44}}{-00.22}{+00.25} & \metric{3.08}{-00.15}{+00.16} \\

                     & $\varphi_{2}$
                     & \metric{0.59}{-0.15}{+0.15} & \metric{0.96}{-0.01}{+0.01} & \metric{0.99}{-0.01}{+0.01} & \metric{0.97}{-0.04}{+0.03} & \metric{\textbf{0.99}}{-0.01}{+0.01}
                     & $\psi_{2}$ & \metric{2.43}{-00.85}{+00.84} & \metric{2.68}{-0.51}{+0.52} & \metric{4.10}{-00.36}{+00.38} & \metric{\textbf{4.55}}{-00.42}{+00.48} \\

                     & $\varphi_{3}$
                     & \metric{0.51}{-0.17}{+0.16} & \metric{0.85}{-0.04}{+0.04} & \metric{0.84}{-0.04}{+0.03} & \metric{0.93}{-0.01}{+0.01} & \metric{\textbf{0.94}}{-0.01}{+0.01}
                     & $\psi_{3}$ & \metric{2.31}{-00.37}{+00.41} & \metric{1.71}{-0.21}{+0.20} & \metric{2.78}{-00.09}{+00.10} & \metric{\textbf{3.19}}{-00.49}{+00.41} \\

                     & $\varphi_{4}$
                     & \metric{0.75}{-0.13}{+0.11} & \metric{0.94}{-0.02}{+0.02} & \metric{0.93}{-0.02}{+0.02} & \metric{\textbf{0.99}}{-0.00}{+0.00} & \metric{0.98}{-0.01}{+0.01}
                     & $\psi_{4}$ & \metric{563.08}{-65.50}{+55.74} & \metric{42.97}{-9.26}{+9.50} & \metric{411.57}{-96.55}{+101.14} & \metric{\textbf{646.26}}{-47.89}{+55.04} \\

                     & $\varphi_{5}$
                     & \metric{0.91}{-0.03}{+0.03} & \metric{0.96}{-0.02}{+0.01} & \metric{0.98}{-0.01}{+0.01} & \metric{\textbf{1.00}}{-0.00}{+0.00} & \metric{0.98}{-0.01}{+0.01}
                     & $\psi_{5}$ & \metric{356.14}{-46.19}{+42.70} & \metric{30.82}{-7.22}{+7.28} & \metric{332.52}{-77.14}{+78.00} & \metric{\textbf{465.48}}{-40.86}{+44.85} \\

                     & $\varphi_{6}$
                     & \metric{0.28}{-0.13}{+0.13} & \metric{0.88}{-0.02}{+0.02} & \metric{0.52}{-0.10}{+0.09} & \metric{0.49}{-0.02}{+0.02} & \metric{\textbf{0.94}}{-0.02}{+0.01}
                     & $\psi_{6}$ & \metric{546.22}{-104.59}{+103.85} & \metric{32.68}{-6.00}{+6.30} & \metric{407.31}{-90.89}{+97.10} & \metric{\textbf{651.15}}{-81.99}{+73.60} \\

                     & $\varphi_{7}$
                     & \metric{0.35}{-0.09}{+0.10} & \metric{0.89}{-0.03}{+0.02} & \metric{0.59}{-0.11}{+0.10} & \metric{0.65}{-0.01}{+0.01} & \metric{\textbf{0.94}}{-0.01}{+0.01}
                     & $\psi_{7}$ & \metric{527.50}{-115.67}{+111.03} & \metric{28.99}{-4.83}{+5.09} & \metric{374.16}{-78.07}{+84.21} & \metric{\textbf{672.39}}{-82.47}{+72.40} \\

                     & $\varphi_{8}$
                     & \metric{0.35}{-0.13}{+0.15} & \metric{0.78}{-0.05}{+0.05} & \metric{0.50}{-0.09}{+0.09} & \metric{0.51}{-0.02}{+0.02} & \metric{\textbf{0.92}}{-0.02}{+0.02}
                     & $\psi_{8}$ & \metric{361.99}{-77.65}{+73.37} & \metric{35.78}{-5.56}{+6.64} & \metric{191.42}{-60.49}{+64.64} & \metric{\textbf{491.31}}{-91.77}{+83.62} \\
                    \midrule
                    \multirow{10}{*}{\rotatebox[origin=c]{90}{WarehouseEnv}}

                     & $\varphi_{9}$
                     & \metric{0.98}{-0.02}{+0.01} & \metric{0.98}{-0.01}{+0.01} & \metric{\textbf{0.99}}{-0.01}{+0.01} & \multicolumn{1}{c}{\textemdash} & \metric{0.99}{-0.01}{+0.01}
                     & $\psi_{9}$ & \metric{2.52}{-00.19}{+00.19} & \metric{2.45}{-00.22}{+00.22} & \multicolumn{1}{c}{\textemdash} & \metric{\textbf{2.78}}{-00.07}{+00.07} \\

                     & $\varphi_{10}$
                     & \metric{0.56}{-0.29}{+0.25} & \metric{0.95}{-0.01}{+0.01} & \metric{0.84}{-0.19}{+0.10} & \multicolumn{1}{c}{\textemdash} & \metric{\textbf{0.95}}{-0.01}{+0.01}
                     & $\psi_{10}$ & \metric{13.84}{-02.73}{+03.15} & \metric{13.21}{-04.65}{+06.05} & \multicolumn{1}{c}{\textemdash} & \metric{\textbf{15.92}}{-06.48}{+07.55} \\

                     & $\varphi_{11}$
                     & \metric{0.46}{-0.25}{+0.26} & \metric{0.99}{-0.02}{+0.01} & \metric{0.91}{-0.13}{+0.08} & \multicolumn{1}{c}{\textemdash} & \metric{\textbf{0.99}}{-0.00}{+0.00}
                     & $\psi_{11}$ & \metric{2.32}{-00.37}{+00.32} & \metric{2.27}{-00.82}{+00.63} & \multicolumn{1}{c}{\textemdash} & \metric{\textbf{3.19}}{-00.17}{+00.17} \\

                     & $\varphi_{12}$
                     & \metric{0.36}{-0.22}{+0.23} & \metric{0.88}{-0.04}{+0.03} & \metric{\textbf{0.93}}{-0.05}{+0.04} & \multicolumn{1}{c}{\textemdash} & \metric{0.80}{-0.09}{+0.08}
                     & $\psi_{12}$ & \metric{2.07}{-00.70}{+00.66} & \metric{2.35}{-00.85}{+00.69} & \multicolumn{1}{c}{\textemdash} & \metric{\textbf{3.93}}{-00.35}{+00.39} \\

                     & $\varphi_{13}$
                     & \metric{0.42}{-0.28}{+0.32} & \metric{0.91}{-0.06}{+0.05} & \metric{0.80}{-0.30}{+0.20} & \multicolumn{1}{c}{\textemdash} & \metric{\textbf{0.99}}{-0.00}{+0.00}
                     & $\psi_{13}$ & \metric{823.49}{-128.61}{+81.49} & \metric{131.88}{-52.17}{+72.50} & \multicolumn{1}{c}{\textemdash} & \metric{\textbf{880.60}}{-35.05}{+30.81} \\

                     & $\varphi_{14}$
                     & \metric{0.62}{-0.32}{+0.28} & \metric{0.58}{-0.24}{+0.23} & \metric{0.78}{-0.29}{+0.20} & \multicolumn{1}{c}{\textemdash} & \metric{\textbf{0.97}}{-0.03}{+0.02}
                     & $\psi_{14}$ & \metric{433.90}{-152.40}{+146.89} & \metric{114.73}{-48.11}{+49.09} & \multicolumn{1}{c}{\textemdash} & \metric{\textbf{656.79}}{-163.11}{+127.06} \\

                     & $\varphi_{15}$
                     & \metric{0.41}{-0.28}{+0.31} & \metric{0.67}{-0.19}{+0.17} & \metric{0.06}{-0.04}{+0.05} & \multicolumn{1}{c}{\textemdash} & \metric{\textbf{0.95}}{-0.06}{+0.03}
                     & $\psi_{15}$ & \metric{650.87}{-196.31}{+187.21} & \metric{10.02}{-06.46}{+08.20} & \multicolumn{1}{c}{\textemdash} & \metric{\textbf{682.62}}{-133.97}{+122.94} \\

                     & $\varphi_{16}$
                     & \metric{0.12}{-0.12}{+0.24} & \metric{0.57}{-0.15}{+0.15} & \metric{0.74}{-0.21}{+0.17} & \multicolumn{1}{c}{\textemdash} & \metric{\textbf{0.84}}{-0.12}{+0.10}
                     & $\psi_{16}$ & \metric{219.22}{-162.27}{+188.46} & \metric{13.98}{-09.35}{+13.27} & \multicolumn{1}{c}{\textemdash} & \metric{\textbf{351.43}}{-123.51}{+115.60} \\

                     & $\varphi_{17}$
                     & \metric{0.15}{-0.14}{+0.23} & \metric{0.66}{-0.21}{+0.19} & \metric{0.71}{-0.24}{+0.20} & \multicolumn{1}{c}{\textemdash} & \metric{\textbf{0.96}}{-0.02}{+0.02}
                     & $\psi_{17}$ & \metric{\textbf{783.72}}{-122.94}{+105.80} & \metric{212.38}{-140.50}{+172.09} & \multicolumn{1}{c}{\textemdash} & \metric{577.07}{-172.36}{+170.65} \\

                     & $\varphi_{18}$
                     & \metric{0.06}{-0.06}{+0.08} & \metric{0.52}{-0.19}{+0.19} & \metric{0.74}{-0.22}{+0.18} & \multicolumn{1}{c}{\textemdash} & \metric{\textbf{0.88}}{-0.18}{+0.10}
                     & $\psi_{18}$ & \metric{575.43}{-177.15}{+174.13} & \metric{113.44}{-101.65}{+139.29} & \multicolumn{1}{c}{\textemdash} & \metric{\textbf{857.86}}{-31.28}{+25.25} \\
                    \bottomrule
                \end{tabular}
            }
        \end{small}
    \end{center}
\end{table*}

\begin{figure}[!t]
  \centering
\begingroup
  \captionof{table}{Average steps to successful completion ($\downarrow$) on finite-horizon tasks, with 95\% bootstrap confidence intervals; protocol as in \mbox{\cref{tab:main-results}}.}
  \label{tab:finite-steps}
  \centering
  \footnotesize
  \newcommand{\stepmetric}[3]{$#1_{\scriptscriptstyle #2}^{\scriptscriptstyle #3}$}

  \newsavebox{\zonestepsbox}
  \newsavebox{\warehousestepsbox}
  \newlength{\zonestepsrowgap}
  \setlength{\zonestepsrowgap}{0pt}
  \newcommand{\zonestepstable}{%
    \begin{tabular}[t]{@{}l rrrrr@{}}
      \toprule
      \multicolumn{6}{c}{\textbf{ZoneEnv-NM}} \\
      \midrule
      $\varphi$ & LTL2Action & DeepLTL & GenZ-LTL$^\dagger$ & GenZ-LTL & StructLTL \\
      \midrule
      $\varphi_{1}$ &\stepmetric{441.6}{-54}{+47} & \stepmetric{275.7}{-24}{+25} & \stepmetric{281.0}{-32}{+29} & \stepmetric{\textbf{178.6}}{-10}{+09} & \stepmetric{222.0}{-10}{+10} \\[\zonestepsrowgap]
      $\varphi_{2}$ &\stepmetric{421.2}{-44}{+40} & \stepmetric{257.6}{-17}{+22} & \stepmetric{252.9}{-26}{+26} & \stepmetric{\textbf{188.2}}{-26}{+30} & \stepmetric{193.5}{-11}{+12} \\[\zonestepsrowgap]
      $\varphi_{3}$ &\stepmetric{334.6}{-74}{+75} & \stepmetric{201.9}{-16}{+17} & \stepmetric{183.2}{-16}{+17} & \stepmetric{\textbf{125.6}}{-06}{+05} & \stepmetric{156.8}{-10}{+10} \\[\zonestepsrowgap]
      $\varphi_{4}$ &\stepmetric{289.9}{-52}{+58} & \stepmetric{186.7}{-16}{+18} & \stepmetric{175.3}{-18}{+17} & \stepmetric{\textbf{121.6}}{-07}{+06} & \stepmetric{139.9}{-07}{+07} \\[\zonestepsrowgap]
      $\varphi_{5}$ &\stepmetric{148.0}{-20}{+20} & \stepmetric{135.5}{-12}{+12} & \stepmetric{126.2}{-11}{+10} & \stepmetric{\textbf{83.9}}{-04}{+04} & \stepmetric{102.8}{-06}{+05} \\[\zonestepsrowgap]
      $\varphi_{6}$ &\stepmetric{450.2}{-59}{+58} & \stepmetric{291.0}{-14}{+15} & \stepmetric{280.6}{-17}{+16} & \stepmetric{\textbf{180.5}}{-13}{+11} & \stepmetric{238.0}{-13}{+14} \\[\zonestepsrowgap]
      $\varphi_{7}$ &\stepmetric{192.0}{-40}{+45} & \stepmetric{179.6}{-15}{+15} & \stepmetric{163.5}{-16}{+16} & \stepmetric{\textbf{111.9}}{-08}{+07} & \stepmetric{125.6}{-05}{+05} \\[\zonestepsrowgap]
      $\varphi_{8}$ &\stepmetric{566.3}{-36}{+38} & \stepmetric{333.5}{-18}{+17} & \stepmetric{338.1}{-29}{+26} & \stepmetric{\textbf{227.0}}{-18}{+16} & \stepmetric{276.1}{-16}{+17} \\
      \bottomrule
    \end{tabular}%
  }
  \sbox{\warehousestepsbox}{%
    \begin{tabular}[t]{@{}l rrrr@{}}
      \toprule
      \multicolumn{5}{c}{\textbf{WarehouseEnv}} \\
      \midrule
      $\varphi$ & LTL2Action & DeepLTL & GenZ-LTL$^\dagger$ & StructLTL \\
      \midrule
      $\varphi_{9}$ &\stepmetric{226.3}{-013}{+013} & \stepmetric{\textbf{196.5}}{-05}{+05} & \stepmetric{221.4}{-08}{+09} & \stepmetric{202.1}{-07}{+08} \\
      $\varphi_{10}$ &\stepmetric{169.7}{-008}{+010} & \stepmetric{163.4}{-16}{+18} & \stepmetric{157.2}{-09}{+11} & \stepmetric{\textbf{143.7}}{-06}{+07} \\
      $\varphi_{11}$ &\stepmetric{346.2}{-097}{+093} & \stepmetric{\textbf{160.2}}{-10}{+13} & \stepmetric{179.2}{-17}{+17} & \stepmetric{173.2}{-19}{+25} \\
      $\varphi_{12}$ &\stepmetric{387.2}{-040}{+046} & \stepmetric{353.1}{-27}{+35} & \stepmetric{\textbf{349.8}}{-18}{+17} & \stepmetric{371.2}{-19}{+21} \\
      $\varphi_{13}$ &\stepmetric{331.3}{-120}{+136} & \stepmetric{185.8}{-28}{+31} & \stepmetric{201.4}{-65}{+89} & \stepmetric{\textbf{126.8}}{-03}{+03} \\
      $\varphi_{14}$ &\stepmetric{304.3}{-079}{+102} & \stepmetric{323.3}{-69}{+81} & \stepmetric{202.0}{-09}{+08} & \stepmetric{\textbf{193.3}}{-10}{+15} \\
      $\varphi_{15}$ &\stepmetric{220.8}{-053}{+059} & \stepmetric{151.3}{-18}{+20} & \stepmetric{210.0}{-62}{+61} & \stepmetric{\textbf{139.5}}{-06}{+06} \\
      $\varphi_{16}$ &\stepmetric{573.2}{-075}{+115} & \stepmetric{430.5}{-52}{+55} & \stepmetric{338.5}{-25}{+30} & \stepmetric{\textbf{329.3}}{-30}{+35} \\
      $\varphi_{17}$ &\stepmetric{378.2}{-064}{+088} & \stepmetric{211.4}{-41}{+54} & \stepmetric{169.0}{-15}{+22} & \stepmetric{\textbf{151.1}}{-04}{+05} \\
      $\varphi_{18}$ &\stepmetric{613.4}{-125}{+095} & \stepmetric{449.4}{-54}{+53} & \stepmetric{339.3}{-33}{+41} & \stepmetric{\textbf{300.0}}{-17}{+23} \\
      \bottomrule
    \end{tabular}%
  }
  \sbox{\zonestepsbox}{\zonestepstable}
  \setlength{\zonestepsrowgap}{\dimexpr
    \ht\warehousestepsbox+\dp\warehousestepsbox
    -\ht\zonestepsbox-\dp\zonestepsbox\relax}
  \divide\zonestepsrowgap by 7
  \resizebox{\linewidth}{!}{%
    \zonestepstable\hspace{12pt}%
    \usebox{\warehousestepsbox}%
  }\par
\endgroup

  \newsavebox{\compositionaltablebox}
  \sbox{\compositionaltablebox}{%
    \begin{minipage}[b]{0.375\textwidth}
      \centering
      {\footnotesize
        \setlength{\tabcolsep}{2pt}
        \newcommand{\metric}[3]{$#1_{\scriptscriptstyle #2}^{\scriptscriptstyle #3}$}
        \begin{tabular*}{\linewidth}{@{\extracolsep{\fill}}l ccc@{}}
          \toprule
          & DeepLTL            & GenZ-LTL$^\dagger$ & StructLTL                   \\
          \midrule
          $\varphi_{19}$ & \metric{0.78}{-0.14}{+0.13} & \metric{0.77}{-0.23}{+0.18} & \metric{\textbf{0.97}}{-0.04}{+0.02} \\
          $\varphi_{20}$ & \metric{0.70}{-0.21}{+0.18} & \metric{0.75}{-0.22}{+0.17} & \metric{\textbf{0.96}}{-0.03}{+0.02} \\
          $\varphi_{21}$ & \metric{0.70}{-0.25}{+0.21} & \metric{0.57}{-0.24}{+0.22} & \metric{\textbf{0.95}}{-0.03}{+0.02} \\
          $\varphi_{22}$ & \metric{0.04}{-0.04}{+0.06} & \metric{0.59}{-0.27}{+0.25} & \metric{\textbf{0.96}}{-0.05}{+0.03} \\
          $\varphi_{23}$ & \metric{0.11}{-0.11}{+0.15} & \metric{0.65}{-0.23}{+0.20} & \metric{\textbf{0.89}}{-0.16}{+0.09} \\
          $\varphi_{24}$ & \metric{0.12}{-0.11}{+0.17} & \metric{0.45}{-0.22}{+0.23} & \metric{\textbf{0.77}}{-0.21}{+0.16} \\
          \bottomrule
        \end{tabular*}
      }
      \captionof{table}{Compositional generalisation results over 10 seeds and 512 episodes per seed, with 95\% bootstrapped confidence intervals.}
      \label{tab:comp_gen}
    \end{minipage}%
  }

  \newsavebox{\scaledcompositionaltablebox}
  \sbox{\scaledcompositionaltablebox}{%
    \resizebox{129.11532pt}{!}{\usebox{\compositionaltablebox}}%
  }

  \vspace{12pt}
  \begin{minipage}[b]{\dimexpr\textwidth-\wd\scaledcompositionaltablebox-12pt\relax}
    \centering
    \includegraphics[width=\linewidth]{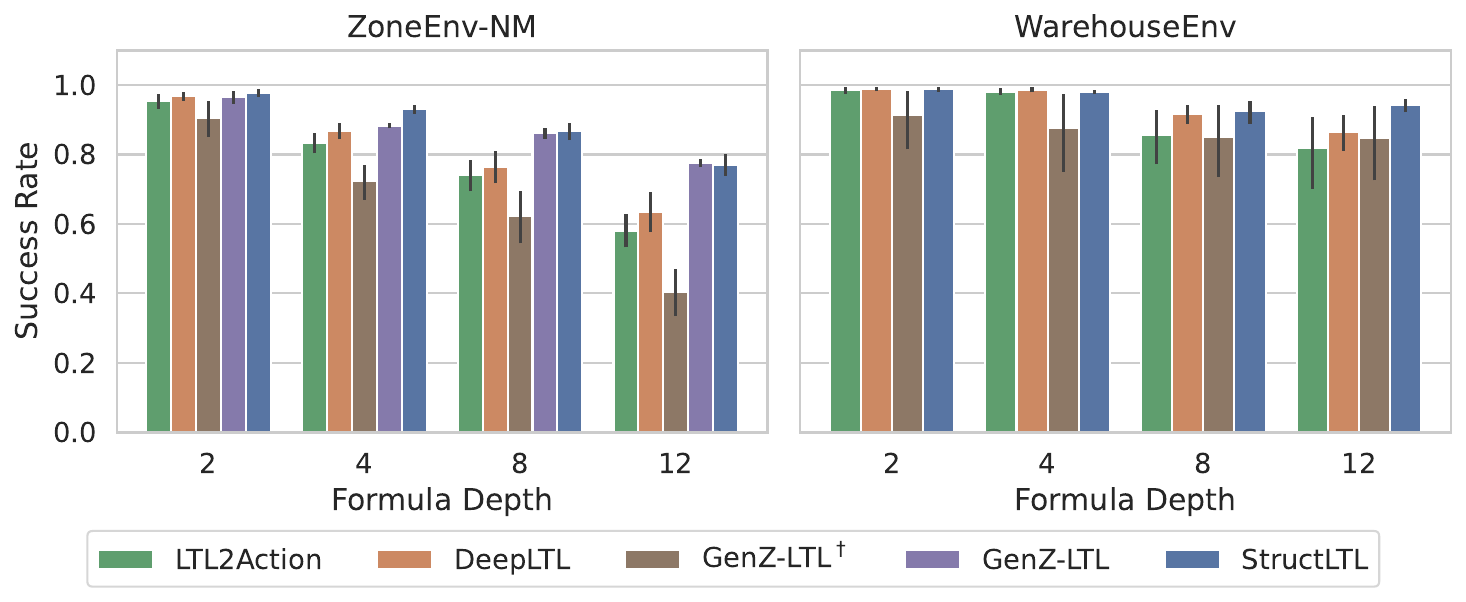}
    \captionof{figure}{Generalisation to reach-tasks of increasing depth.}
    \label{fig:gen_depth}
  \end{minipage}\hspace{12pt}%
  \usebox{\scaledcompositionaltablebox}
\end{figure}

\paragraph {Evaluation protocol.}
DeepLTL and StructLTL are trained with the same generic training curriculum (see \cref{app:curriculum} for further details).
LTL2Action originally does not employ curriculum training, but we found this to be beneficial and thus use an adapted training curriculum for a fair comparison.
We train using PPO~\citep{schulman2017Proximal}, with detailed hyperparameters in \cref{app:hyperparameters}.
GenZ-LTL instead uses a safety-aware PPO procedure~\citep{yu2022reachability} and is trained by sampling feasible reach-avoid subgoals.
We follow standard practice~\citep{qiu2023Instructing,jackermeier2025deepltl,guo2025One} and report success rates and number of steps until successful completion for finite-horizon tasks, and the number of successfully completed accepting cycles over a fixed evaluation horizon of 1000 steps for infinite-horizon tasks.
Note that LTL2Action is not applicable to infinite-horizon tasks due to its reliance on LTL progression, which never yields positive reward signals for non-co-safe formulae.
All results are averaged over 10 random seeds and 512 episodes per seed, and we provide 95\% stratified bootstrap confidence intervals~\citep{agarwal2021Deep}.

\subsection{Results}
\begin{figure}[!t]
  \centering
  \includegraphics[width=0.8\linewidth]{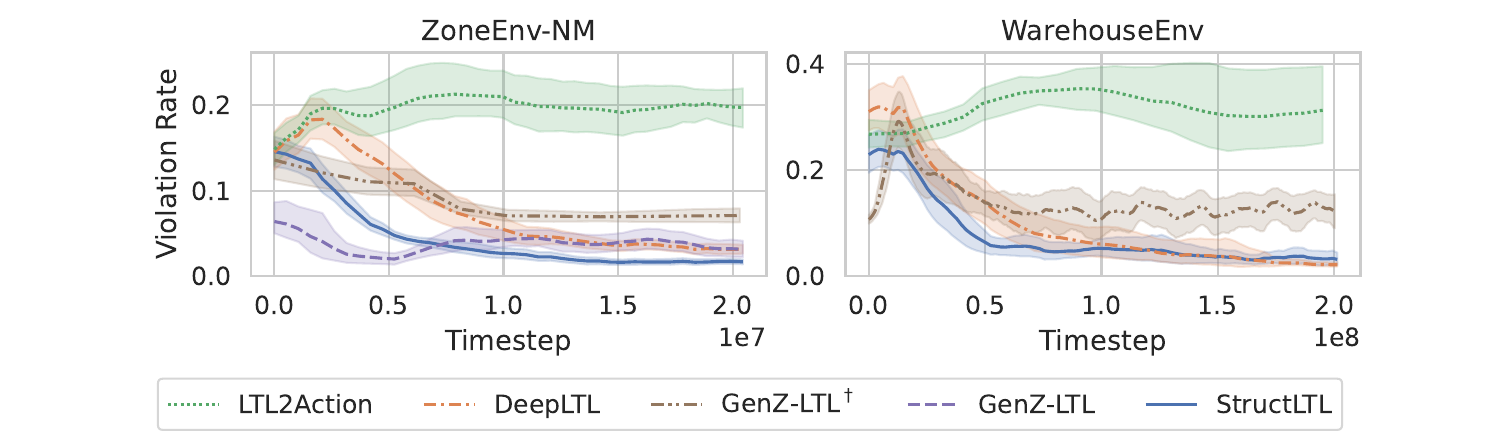}
  \caption{Violation rates over training on tasks that can be violated per environment (50 episodes each). Shading shows 95\% confidence intervals across seeds.}
  \label{fig:violations}
  \vspace{-12pt}
\end{figure}

\cref{fig:curves} shows evaluation curves averaged over finite-horizon tasks over training, and \cref{tab:main-results} lists results of the trained policies on both finite- and infinite-horizon specifications. \Cref{tab:finite-steps} reports the number of steps until successful completion of finite-horizon tasks, and \cref{fig:violations} shows violation rates over training. We visualise trajectories produced by StructLTL in \cref{fig:zones-example-traj,fig:warehouse-example-traj,app:trajectories}.

StructLTL consistently achieves high success rates in finite-horizon tasks, in most cases exceeding~$90\%$.
This is true not only for the simpler specifications in ZoneEnv-NM, but also for complex tasks requiring compositional reasoning in the Warehouse environment (\textbf{Q1}).
In comparison to the baselines, we overall achieve significantly higher performance,
particularly on tasks requiring non-myopic reasoning ($\varphi_{6}$\,--\,$\varphi_{8}$) (\textbf{Q2}) and involving complex combinations of propositions ($\varphi_{14}$\,--\,$\varphi_{18}$).
Violation rates are also generally lower than for the baselines.
These results highlight the benefits of our structured representations: whereas our approach represents complex combinations of atomic propositions via succinct Boolean formulae, DeepLTL and GenZ-LTL struggle to handle the combinatorial explosion of assignments.
LTL2Action underperforms even on simpler tasks, consistent with previously reported results~\cite{jackermeier2025deepltl,guo2025One} (\textbf{Q3}).

Notably, StructLTL achieves similar success rates to GenZ-LTL in ZoneEnv-NM, even when the latter employs a hand-crafted, environment-specific observation reduction function.
Without this observation reduction its performance notably degrades (GenZ-LTL$^\dagger$), and our method thus performs significantly better.
However, we observe that the reduced observation space\,---\,where applicable\,---\,allows GenZ-LTL to learn more efficient policies, requiring fewer steps until task completion.
In WarehouseEnv, where GenZ-LTL's observation reduction is not applicable, StructLTL generally achieves significantly higher success rates while simultaneously requiring fewer steps.
The results on infinite-horizon tasks paint a similar picture:
our approach consistently completes more accepting cycles in the LDBA for almost all formulae, suggesting that it is both better and more efficient at achieving challenging infinite-horizon specifications.

\paragraph{Generalisation study.}
We investigate the generalisation ability of our approach to long Boolean formulae sequences.
During training, we sample sequences of maximum length $2$.
\cref{fig:gen_depth} shows success rates obtained on randomly sampled nested LTL formulae $\mathsf F(\psi_1\land\mathsf F(\psi_2\land\ldots))$ of depth up to 12.
While longer formulae sequences are naturally more challenging, we observe that StructLTL achieves high success rates on these out-of-distribution tasks and performs better than the baselines.

We furthermore conduct a compositional generalisation experiment: we train baseline methods and StructLTL in WarehouseEnv with a maximum conjunction length of 2, and test on tasks involving unseen conjunctions of length 3.
Task details are provided in \cref{app:tasks}.
The results in \cref{tab:comp_gen} show that StructLTL successfully generalises to tasks involving unseen Boolean formulae.


\paragraph{Further results and ablations.}
In \cref{app:nm} we further validate the non-myopic reasoning capabilities of our approach. Furthermore, we conduct ablation studies in \cref{app:ablation} to analyse the impact of the hierarchical DNF encoder (comparing to GNN and RNN variants) and the temporal attention mechanism.
We find that both components significantly contribute to StructLTL's performance.
We also ablate the impact of curriculum learning (\cref{app:curriculum-learning}), demonstrating that it leads to improved sample efficiency and performance. At the same time, the results show that our method is also able to learn successfully with no curriculum at all.
We analyse the computational complexity and practical runtime of our method in \cref{app:complexity}.



\section{Conclusion, Limitations \& Future Work}
\label{sec:conclusion_limitations}

We presented StructLTL, a novel approach for learning generalist policies to follow LTL instructions based on structured task representations. Our method exploits the structure of B\"uchi automata constructed from a given specification, extracting sequences of Boolean formulae that succinctly represent different ways of achieving the task. These formulae are encoded with a specialised hierarchical neural network architecture, and we furthermore introduce a temporal attention mechanism to enable non-myopic reasoning about future subgoals. In contrast to previous methods, our structured representations effectively capture interactions between atomic propositions, thereby enabling compositional generalisation. We demonstrated that our approach can effectively zero-shot generalise to unseen LTL instructions and outperforms state-of-the-art methods across complex domains and tasks.

A limitation of our method is that\,---\,like previous approaches\,---\,we rely on ground-truth access to the labelling function $L$ (which maps from observations to
atomic propositions).
We are excited to explore ideas such as jointly learning the labelling function
from visual or sensory input, or leveraging pre-trained foundation models as high-level event detectors in the future, as investigated in recent work~\citep{li2025GroundComposeReinforce,parac2026RMVLM}.
Another interesting direction for future work is to incorporate techniques such as hindsight experience replay \citep{andrychowicz2017Hindsight,voloshin2023Eventual} to improve the sample efficiency and performance of our method.
We believe that future advances in these areas could directly lead to real-world applications, such as in autonomous robotic domains.

\clearpage

\section*{Acknowledgments}
This work was supported by the UKRI Erlangen AI Hub on Mathematical and Computational Foundations of AI (No. EP/Y028872/1). MJ and JC are funded by the EPSRC Centre for Doctoral Training in Autonomous Intelligent Machines and Systems (EP/S024050/1).


\bibliographystyle{plainnat}  
\bibliography{rl,mc,ml}


\clearpage
\appendix
\crefalias{section}{appendix}
\crefalias{subsection}{appendix}

\section{Semantics of LTL}
\label{app:ltl_semantics}

The satisfaction semantics of LTL are specified by the satisfaction relation $\sigma\models\varphi$, which is recursively defined as follows~\citep{pnueli1977temporal}:
\begin{align*}
  \sigma & \models \top                                                                                                                                                   \\
  \sigma & \models \mathsf a                 &  & \text{iff } \mathsf a\in \sigma_0                                                                                       \\
  \sigma & \models \varphi\land\psi          &  & \text{iff } \sigma\models\varphi \land \sigma\models\psi                                                                \\
  \sigma & \models \neg\varphi               &  & \text{iff } \sigma\not\models\varphi                                                                                    \\
  \sigma & \models \nex\varphi               &  & \text{iff } \sigma_{1\ldots}\models\varphi                                                                              \\
  \sigma & \models \varphi\;\mathsf{U}\;\psi &  & \text{iff } \exists j\geq 0.\; \sigma_{j\ldots}\models\psi \land \forall 0\leq i < j.\; \sigma_{i\ldots}\models\varphi.
\end{align*}

\section{Algorithm to Identify Accepting Runs}
\label{app:alg}
\cref{alg:cycles} presents the DFS-based procedure from \citet{jackermeier2025deepltl} for computing paths from a given LDBA state to accepting cycles.
This yields a set of accepting runs that we then convert to sequences of Boolean formulae.
We employ two practical optimisations during this conversion process:
first, as observed by \citet{jackermeier2025deepltl}, requiring the policy to
avoid \emph{all} other LDBA states $q\not\in \{q_i,q_{i+1}\}$ at step $i$ can be
overly restrictive. Instead, we construct $\beta_i^-$ to
only avoid assignments that lead to: (i) sink states (states from which satisfying
the specification is impossible), or (ii) previous states on the current path
(which would undo progress). This relaxation allows the policy more flexibility
while still preventing catastrophic transitions.
Second, when a positive formula $\beta_i^+$ contains a disjunction, we generate a separate formula sequence for each disjunct in $\beta_i^+$.
This avoids requiring the policy to internally select between the different options.
Instead, we employ the value function to choose the approximately optimal formula sequences as described in \cref{sec:test-time}.

\section{Experimental Details}
\label{app:experimental_details}

\subsection{Implementation}
\label{app:implementation}
Our implementation is built in JAX \cite{bradbury_jax_2018}. We use Rabinizer 4 \cite{chockler_rabinizer_2018} to convert LTL specifications into LDBAs. All experiments were run on an NVIDIA GeForce RTX 4070 GPU; we were able to train 10 seeds of a given algorithm in parallel to completion within roughly 30 minutes for ZoneEnv-NM, and roughly 3 hours in Warehouse.

\subsection{Environments}
\label{app:environments}

\paragraph{ZoneEnv-NM.}
The ZoneEnv environment was introduced by \citet{vaezipoor2021LTL2Action}. It consists of a walled plane with 8 circular regions (``zones'') that have four different colours and form the atomic propositions. Since the zones never overlap, the assignments all correspond to singleton sets of atomic propositions (or the empty set). It features a 2-dimensional continuous action space for acceleration and steering. The environment features a high-dimensional state space based on lidar information about the zones, and data from other sensors. Both the zone and robot positions are randomly sampled at the beginning of each episode. If the agent at any point touches a wall, the episode is terminated prematurely. To increase the complexity of the environment, we introduce ZoneEnv-NM, which expands upon ZoneEnv to introduce state-modifying behaviour; when the agent first enters a purple zone, all green zones disappear for the rest of the episode. The agent must learn to reason about this behaviour in order to successfully complete complex tasks. A visualisation of the ZoneEnv-NM environment is provided in \cref{fig:zones}.

\paragraph{Warehouse.}
The warehouse environment consists of a continuous state space, on which a point agent moves via continuous actions. The agent can interact with two types of objects: crates and vases. Both the agent position and the object positions are randomly sampled at the beginning of each episode. There are furthermore three different fixed regions in the grid: region A (red), region B (green), and door (purple). The agent observes objects via a lidar sensor, and additionally receives directional information about the regions. The action space is composite: at each step, the agent can apply both a continuous movement action and pick up or drop a crate or vase. The atomic propositions specify whether the agent is currently carrying a vase or a crate, and if the agent is in one of the three regions. \cref{tab:assignments} lists all possible combinations of propositions that hold true at some state in the Warehouse environment. See \cref{fig:warehouse-env} for a visualisation of the environment.

\begin{algorithm}[t]
  \caption{Computing paths to accepting cycles~\citep{jackermeier2025deepltl}}
  \label{alg:cycles}
  \begin{algorithmic}[1]
    \REQUIRE LDBA $B = (\mathcal Q, q_0, \Sigma, \delta, \mathcal E, \mathcal F)$ and current state $q$.

    \STATE {\bfseries function} \textsc{DFS}($q, p, i$) \COMMENT{$i$ is index of last seen accepting state, or $-1$}
    \STATE $P \gets \emptyset$
    \IF{$q \in \mathcal{F}$}
    \STATE $i \gets |p|$
    \ENDIF
    \FORALL{$a \in \Sigma \cup \{\varepsilon\}$}
    \STATE $p' \gets [p, q]$
    \STATE $q' \gets \delta(q, a)$
    \IF{$q' \in p$}
    \IF{index of $q'$ in $p \le i$}
    \STATE $P \gets P \cup \{p'\}$
    \ENDIF
    \ELSE
    \STATE $P \gets P \cup \text{\textsc{DFS}}(q', p', i)$
    \ENDIF
    \ENDFOR
    \STATE \textbf{return} $P$
    \STATE {\bfseries end function}

    \STATE \COMMENT{Main entry point}
    \IF{$q \in \mathcal{F}$}
    \STATE $i \gets 0$
    \ELSE
    \STATE $i \gets -1$
    \ENDIF
    \STATE \textbf{return} \textsc{DFS}($q, [\,], i$)
  \end{algorithmic}
\end{algorithm}

\begin{table}[t]
  \centering
  \caption{Possible assignments in the Warehouse environment.}
  \begin{tabular}{@{}ccc@{}}
    \toprule
    $\{\mathsf{region\_a}\}$                                & $\{\mathsf{region\_b}\}$                                & $\{\mathsf{door}\}$                                \\
    $\{\mathsf{vase}\}$                                     & $\{\mathsf{crate}\}$                                    &

    $\{\mathsf{region\_a}, \mathsf{vase}\}$                                                                                                                                \\
    $\{\mathsf{region\_a}, \mathsf{crate}\}$                & $\{\mathsf{region\_b}, \mathsf{vase}\}$                 & $\{\mathsf{region\_b}, \mathsf{crate}\}$           \\
    $\{\mathsf{door}, \mathsf{vase}\}$                      & $\{\mathsf{door}, \mathsf{crate}\}$                     & $\{\mathsf{vase}, \mathsf{crate}\}$                \\

    $\{\mathsf{region\_a}, \mathsf{vase}, \mathsf{crate}\}$ & $\{\mathsf{region\_b}, \mathsf{vase}, \mathsf{crate}\}$ & $\{\mathsf{door}, \mathsf{vase}, \mathsf{crate}\}$ \\

                                                            & $\emptyset$                                             &                                                    \\
    \bottomrule
  \end{tabular}
  \label{tab:assignments}
\end{table}

\subsection{Training Curricula}
\label{app:curriculum}
We develop a generic training curriculum for each environment, detailed in \cref{tab:training_curricula}. Each curriculum stage consists of a distribution over sequences of Boolean formulae in DNF\@. Later curriculum stages generally consist of more difficult tasks. Once the agent has achieved sufficient success on tasks sampled from the current stage (as measured by successes over the last 256 episodes), we move on to the next stage.
We define four general, environment-independent sets of formulae used to construct the curricula:
\textsc{props} are simply formulae of the form $p$, where $p\in AP$. \textsc{And} consists of conjunctions of propositions up to length 3. \textsc{AndNot} consists of formulae of the form $p_1\land\neg p_2$. Finally, \textsc{Or} consists of disjunctions of all possible formulae from the previous sets.
We furthermore distinguish two sequence types: \textsc{Reach-Avoid} sequences of the form $((\beta_1^+,\beta_1^-),\ldots,(\beta_n^+,\beta_n^-))$ and \textsc{Reach-Stay} sequences $((\beta_\varepsilon,\beta_1^-),(\varphi,\neg\varphi),(\varphi,\neg\varphi),\ldots)$ that contain the $\varepsilon$-token $\beta_\varepsilon$ and require the agent to at some point hold formula $\varphi$ for $n$ time steps.
For DeepLTL we compute satisfying assignments for each formula in order to obtain a sequence of assignments, rather than a sequence of formulae.

LTL2Action does not originally employ curriculum learning, but we experimented with various training curricula for a fair comparison.
We found it important to include diverse LTL formulae even in early curriculum stages (in contrast to the curricula for the other methods), since LTL2Action operates directly on the syntax level and hence requires exposure to all the various LTL operators.
In ZoneEnv-NM, we sample formulae of the reach-avoid family of the form $\neg \mathsf{a}\until (\mathsf (\mathsf b\lor\mathsf c) \land \ldots)$ with increasing depth and number of disjunctions in later curriculum stages.
We use the same family for the Warehouse curriculum, but furthermore sample Boolean formulae from the \textsc{And} and \textsc{AndNot} families instead of simple atomic propositions in later stages.
Note that LTL2Action does not support infinite-horizon specifications, and we hence do not sample formulae containing the $\always$ operator.

\begin{table*}[t]
  \caption{Curricula for training sequences. For each stage, we list the success rate threshold $\kappa$ for progression  (measured over a rolling window of the $256$ latest episodes), the sequence type (reach-avoid/reach-stay), the probability $\mathrm{P}_\mathrm{seq}$ of sampling a sequence of that type, the number of steps $n$ in the sequence (or, for reach-stay tasks, the number of consecutive time steps for satisfying the ``stay'' sub-task), the type of each reach ($\beta_i^+$) and avoid ($\beta_i^-$) Boolean formula, and the probability $\mathrm{P}_\mathrm{avoid}$ of sampling avoid formulae at each timestep. Dashes (---) indicate ``not applicable''.}

  \label{tab:training_curricula}
  \begin{center}
    \begin{small}

      \begin{tabular}{llll ccccc}
        \toprule

         & Stage              & $\kappa$             & Type        & $\mathrm{P}_\mathrm{seq}$ & $n$    & $\beta_i^+$     & $\beta_i^-$    & $\mathrm{P}_\mathrm{avoid}$ \\

        \midrule

        \multirow{12}{*}{\rotatebox[origin=c]{90}{ZoneEnv-NM}}

         & 1                  & 0.9                  & Reach-Avoid & 1.0                       & 1      & props           & ---            & 0.0                         \\

        \cmidrule(lr){2-9}

         & 2                  & 0.95                 & Reach-Avoid & 1.0                       & 2      & props           & ---            & 0.0                         \\

        \cmidrule(lr){2-9}

         & 3                  & 0.95                 & Reach-Avoid & 1.0                       & 1      & props           & props          & 1.0                         \\

        \cmidrule(lr){2-9}

         & 4                  & 0.9                  & Reach-Avoid & 1.0                       & 2      & props           & props          & 1.0                         \\

        \cmidrule(lr){2-9}

         & \multirow{2}{*}{5} & \multirow{2}{*}{0.9} & Reach-Avoid & 0.4                       & [1, 2] & props           & props+ORs      & 0.7                         \\

         &                    &                      & Reach-Stay  & 0.6                       & 30     & props           & props          & 0.5                         \\

        \cmidrule(lr){2-9}

         & \multirow{2}{*}{6} & \multirow{2}{*}{0.9} & Reach-Avoid & 0.8                       & [1, 2] & props           & props+ORs      & 0.7                         \\

         &                    &                      & Reach-Stay  & 0.2                       & 60     & props           & props          & 0.5                         \\

        \cmidrule(lr){2-9}

         & \multirow{2}{*}{7} & \multirow{2}{*}{---} & Reach-Avoid & 0.8                       & [1, 2] & props           & props+ORs      & 0.7                         \\

         &                    &                      & Reach-Stay  & 0.2                       & 60     & props           & props+ORs      & 0.5                         \\

        \midrule

        \multirow{12}{*}{\rotatebox[origin=c]{90}{Warehouse}}

         & 1                  & 0.9                  & Reach-Avoid & 1.0                       & 1      & props           & ---            & 0.0                         \\

        \cmidrule(lr){2-9}

         & 2                  & 0.95                 & Reach-Avoid & 1.0                       & 1      & all except ORs  & ---            & 0.0                         \\

        \cmidrule(lr){2-9}

         & 3                  & 0.9                  & Reach-Avoid & 1.0                       & 2      & all except ORs  & ---            & 0.0                         \\

        \cmidrule(lr){2-9}

         & 4                  & 0.95                 & Reach-Avoid & 1.0                       & 1      & all except ORs  & all except ORs & 0.5                         \\

        \cmidrule(lr){2-9}

         & 5                  & 0.95                 & Reach-Avoid & 1.0                       & [1,2]  & all except ORs  & all except ORs & 0.5                         \\

        \cmidrule(lr){2-9}

         & \multirow{2}{*}{6} & \multirow{2}{*}{0.9} & Reach-Avoid & 0.4                       & [1, 2] & all except ORs  & all            & 0.5                         \\

         &                    &                      & Reach-Stay  & 0.6                       & 30     & all excepts ORs & all except ORs & 0.2                         \\

        \cmidrule(lr){2-9}

         & \multirow{2}{*}{7} & \multirow{2}{*}{---} & Reach-Avoid & 0.8                       & [1, 2] & all except ORs  & all            & 0.5                         \\

         &                    &                      & Reach-Stay  & 0.2                       & 60     & all except ORs  & all            & 0.5                         \\

        \bottomrule
      \end{tabular}
    \end{small}
  \end{center}
  \vskip -0.1in
\end{table*}

\subsection{Hyperparameters}
\label{app:hyperparameters}
Table~\ref{tab:hyperparameters} lists hyperparameters of StructLTL and the baselines. We report PPO parameters and details on the neural network architectures used for processing environment observations and the actor and critic. ZoneEnv-NM hyperparameters were chosen to match those of ZoneEnv in previous work~\citep{vaezipoor2021LTL2Action,jackermeier2025deepltl} and adapted for the Warehouse environment.
We use the Adam optimiser~\citep{kingma2015Adam} throughout.

\begin{table*}[t]
\centering
\caption{Hyperparameters for {LTL2Action}, {DeepLTL}, {StructLTL}, and {GenZ-LTL} across ZoneEnv-NM and Warehouse. Values spanning multiple columns indicate identical settings across those methods/environments. GenZ-LTL uses a Safe PPO variant, which introduces additional hyperparameters that are not applicable (N/A) to the other methods.}
\label{tab:hyperparameters}
\resizebox{\textwidth}{!}{%
\begin{tabular}{llcccccccc}
\toprule
\multirow{2}{*}{Category} & \multirow{2}{*}{Hyperparameter} & \multicolumn{4}{c}{ZoneEnv-NM} & \multicolumn{4}{c}{WarehouseEnv} \\
\cmidrule(lr){3-6} \cmidrule(lr){7-10}
 &  & {LTL2Action} & {DeepLTL} & {StructLTL} & {GenZ-LTL} & {LTL2Action} & {DeepLTL} & {StructLTL} & {GenZ-LTL} \\
\midrule
\multirow{13}{*}{PPO} 
& Total timesteps & \multicolumn{4}{c}{--- $2 \times 10^7$ ---} & \multicolumn{4}{c}{--- $2 \times 10^8$ ---} \\
& Environments & \multicolumn{4}{c}{--- 16 ---} & 64 & \multicolumn{3}{c}{--- 128 ---} \\
& Steps/update & \multicolumn{4}{c}{--- 4096 ---} & \multicolumn{4}{c}{--- 1024 ---} \\
& Minibatches & \multicolumn{4}{c}{--- 32 ---} & \multicolumn{4}{c}{--- 4 ---} \\
& Update epochs & \multicolumn{4}{c}{--- 10 ---} & \multicolumn{4}{c}{--- 5 ---} \\
& Discount ($\gamma$) & \multicolumn{8}{c}{--- 0.998 ---} \\
& GAE lambda ($\lambda$) & \multicolumn{8}{c}{--- 0.95 ---} \\
& Clip epsilon & \multicolumn{8}{c}{--- 0.2 ---} \\
& Entropy coef. & \multicolumn{8}{c}{--- 0.003 ---} \\
& Value func.\ coef. & \multicolumn{4}{c}{--- 1.0 ---} & \multicolumn{3}{c}{--- 0.5 ---} & 1.0 \\
& Learning rate & \multicolumn{4}{c}{--- $3 \times 10^{-4}$ ---} & \multicolumn{4}{c}{--- $5 \times 10^{-4}$ ---} \\
& Max grad norm & \multicolumn{4}{c}{--- 0.5 ---} & \multicolumn{4}{c}{--- 1.0 ---} \\
& Adam epsilon & \multicolumn{8}{c}{--- $10^{-8}$ ---} \\
\midrule
\multirow{6}{*}{Safe PPO}
& Cost discount ($\gamma_c$) & \multicolumn{3}{c}{--- N/A ---} & 0.998 & \multicolumn{3}{c}{--- N/A ---} & 0.998 \\
& Cost value func.\ coef. & \multicolumn{3}{c}{--- N/A ---} & 1.0 & \multicolumn{3}{c}{--- N/A ---} & 1.0 \\
& Lagrangian coef. & \multicolumn{3}{c}{--- N/A ---} & 1.0 & \multicolumn{3}{c}{--- N/A ---} & 1.0 \\
& Target cost & \multicolumn{3}{c}{--- N/A ---} & $-0.05$ & \multicolumn{3}{c}{--- N/A ---} & $-0.1$ \\
& Min Lagrangian & \multicolumn{3}{c}{--- N/A ---} & $10^{-2}$ & \multicolumn{3}{c}{--- N/A ---} & $10^{-3}$ \\
& Max Lagrangian & \multicolumn{3}{c}{--- N/A ---} & 5.0 & \multicolumn{3}{c}{--- N/A ---} & 3.0 \\
\midrule
Curriculum & Episode window & \multicolumn{8}{c}{--- 256 ---} \\
\midrule
\multirow{3}{*}{Env Net} 
& Hidden sizes & \multicolumn{8}{c}{--- [128] ---} \\
& Output size & \multicolumn{8}{c}{--- 64 ---} \\
& Activation & \multicolumn{8}{c}{--- Tanh ---} \\
\midrule
\multirow{2}{*}{Actor} 
& Hidden sizes & \multicolumn{8}{c}{--- [64, 64, 64] ---} \\
& Activation & \multicolumn{8}{c}{--- ReLU ---} \\
\midrule
\multirow{2}{*}{Critic} 
& Hidden sizes & \multicolumn{8}{c}{--- [64, 64] ---} \\
& Activation & \multicolumn{8}{c}{--- Tanh ---} \\
\midrule
Environment & Max episode length & \multicolumn{8}{c}{--- 1,000 ---} \\
\bottomrule
\end{tabular}
}
\end{table*}

\paragraph{Model architectures.}
All methods employ the same overall model architecture. Environment observations are encoded with an MLP and the current LTL task is encoded in a method-dependent way (see below). The resulting latent representations are then concatenated and fed through actor and critic MLPs, respectively. The critic directly outputs a scalar estimate of the value function, whereas the actor outputs the parameters of the action distribution. In ZoneEnv-NM, the actor consists of an MLP with two output heads for the mean and standard deviation of a diagonal Gaussian. In Warehouse, the actor MLP additionally produces logits for a categorical distribution over the discrete actions. Finally, for StructLTL and DeepLTL the actor MLP additionally has a head that outputs the log-probability of taking an $\varepsilon$-action.

\paragraph{LTL encoding networks.}
LTL2Action encodes the LTL formula using a relational graph convolutional network (RGCN)~\citep{schlichtkrull2018Modeling} applied to the formula's syntax tree.
Following the original paper, it consists of $8$ layers with tanh activations, using an embedding dimension of $32$ for ZoneEnv-NM and $16$ for the Warehouse environment.
The weights of all layers are shared, and each layer receives as input the output from the previous layer concatenated with the original one-hot encodings of the symbols in the syntax tree.

DeepLTL encodes sequences of sets of assignments using $16$-dimensional embeddings (one for each assignment) processed by a DeepSets module~\citep{zaheer2017Deep} with a hidden layer size of $32$ and an output size of $16$, using ReLU activations and summation as the aggregation mechanism.
The encoded reach and avoid sets are then concatenated and processed with a gated recurrent unit (GRU)~\citep{cho2014Properties} with embedding size 32. The GRU is applied back-to-front, and the final hidden state forms the sequence embedding. We use $k=2$ (the number of repeating loops at test time) throughout.

StructLTL uses $16$-dimensional proposition embeddings and processes clauses and disjuncts with separate DeepSets networks, both with no hidden layers, output size $16$, and ReLU activations.
The attention mechanism uses hidden dimension $d_k=32$ and ALiBi slope $m = 0.5$ in ZoneEnv-NM, and $m = 5$ in Warehouse. As in DeepLTL, we set the number of accepting cycles before truncation to $2$.

GenZ-LTL does not directly use an LTL encoding network. Instead, it employs an \textit{observation reduction} mechanism that modifies the given observation $s_t$ based on the current reach-avoid subgoal extracted from the LTL specification. In ZoneEnv-NM, this mechanism produces $37$-dimensional features (proprioceptive information and reach/avoid lidars).
However, observation reduction cannot be applied in Warehouse, since it relies on two assumptions: (i) observations must decompose into \textit{identical} proposition-specific components, and this decomposition must be manually known and specified a priori, and (ii) a suitable, manually specified fusion operator is required to handle multiple avoid subgoals.
Since neither of these assumptions hold in Warehouse, GenZ-LTL falls back to encoding reach-avoid subgoals using a one-hot bitvector representation (see \citep[Section 4.1]{guo2025One}).
The modified observations are then processed by the environment module and rest of the network architecture as in the other methods. We denote this variant as GenZ-LTL$^\dagger$.

\subsection{Evaluation Tasks}
\label{app:tasks}
\cref{tab:ltl_specs} lists the complete LTL specifications used in our evaluation. The ZoneEnv-NM tasks are adapted from established specifications for ZoneEnv from the literature~\citep{jackermeier2025deepltl}, and designed with the non-myopic behavior of ZoneEnv-NM in mind.
Warehouse tasks were designed to encompass a broad range of interesting behaviour (such as transporting objects to different regions) and LTL structures.
Furthermore, \cref{tab:gen_formulae} lists the evaluation specifications used in the compositional generalisation experiment.

\begin{table}[]
  \newcommand{\metric}[3]{%
    $#1_{\scriptscriptstyle #2}^{\scriptscriptstyle #3}$%
  }
  \centering
  \caption{Non-myopic reasoning experiment.}
  \label{tab:nm}
  \resizebox{.9\textwidth}{!}{%
    \begin{tabular}{l ccc}
      \toprule
      Formula                                                                                                                                 & {DeepLTL}                   & {GenZ-LTL}                  & {StructLTL}                          \\
      \midrule
      $\event (\mathsf{purple}) \wedge \event (\mathsf{green})$                                                                               & \metric{0.90}{-0.05}{+0.05} & \metric{0.51}{-0.02}{+0.02} & \metric{\textbf{0.95}}{-0.02}{+0.02} \\
      $\event ((\mathsf{purple} \vee \mathsf{red}) \wedge \event \mathsf{green})$                                                             & \metric{0.77}{-0.23}{+0.23} & \metric{0.49}{-0.03}{+0.03} & \metric{\textbf{0.95}}{-0.02}{+0.02} \\
      $\event \mathsf{purple} \wedge \event (\mathsf{red} \wedge (\event (\mathsf{yellow} \wedge (\neg\mathsf{red} \until \mathsf{green}))))$ & \metric{0.47}{-0.23}{+0.23} & \metric{0.13}{-0.02}{+0.02} & \metric{\textbf{0.81}}{-0.08}{+0.08} \\
      $\neg\mathsf{red} \until (\mathsf{yellow} \wedge \event \mathsf{green}) \wedge \event \mathsf{purple}$                                  & \metric{0.58}{-0.22}{+0.22} & \metric{0.25}{-0.03}{+0.03} & \metric{\textbf{0.84}}{-0.05}{+0.05} \\
      \bottomrule
    \end{tabular}%
  }
\end{table}

\section{Additional Results}

\subsection{Non-myopic Reasoning Experiment}
\label{app:nm}
We further validate the non-myopic reasoning capabilities of StructLTL.
We investigate the formulae listed in \cref{tab:nm} in ZoneEnv-NM\@.
Note that all of these formulae require the agent to visit both $\mathsf{purple}$ and $\mathsf{green}$; the policy hence has to non-myopically reason about the entire specification to determine that it \textit{first} needs to visit $\mathsf{green}$ before visiting $\mathsf{purple}$.
Subgoal-decomposition based approaches such as GenZ-LTL are incapable of performing such reasoning.

The results in \cref{tab:nm} clearly show that StructLTL is able to reason non-myopically about the entire task.
For example, when attempting to solve $\event ((\mathsf{purple}\lor\mathsf{red})\land\event \mathsf{green})$, the policy realises that it has to achieve $\mathsf{green}$ at a later step, and hence tries to avoid $\mathsf{purple}$ in the first step (in ~95\% of cases).
As expected, StructLTL performs much better than the myopic baseline GenZ-LTL on these tasks, and we also observe that it outperforms DeepLTL\@.
These results further validate the design of our temporal attention mechanism.

\subsection{Ablation Studies}
\label{app:ablation}

\begin{figure*}
  \centering
  \begin{subfigure}{0.49\columnwidth}
    \centering
    \includegraphics[width=\textwidth]{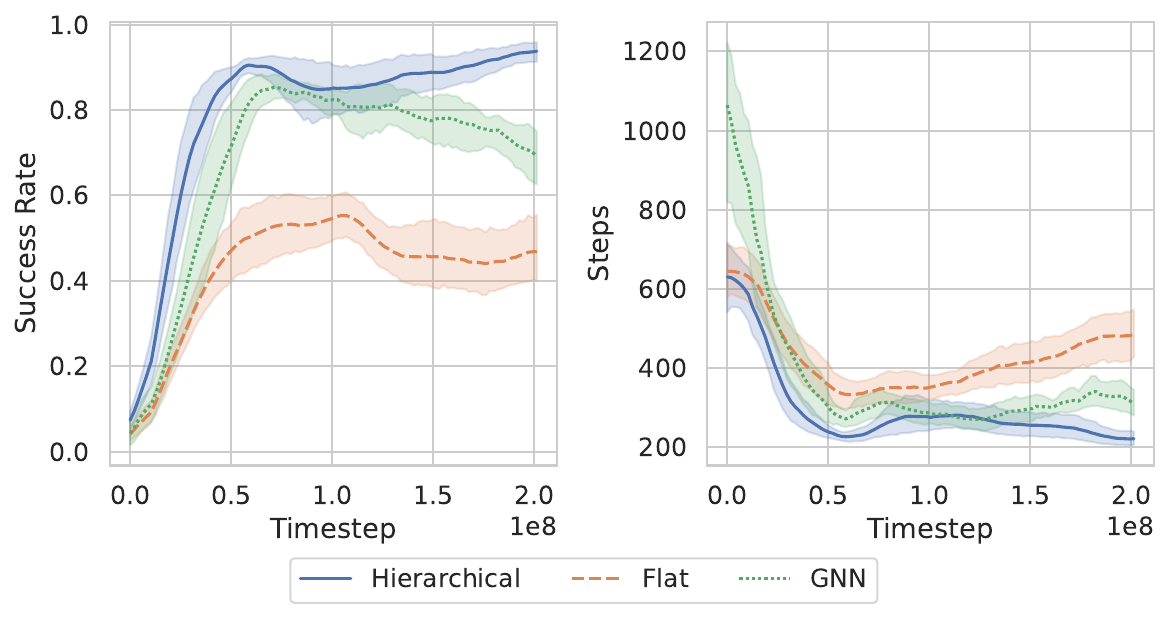}
    \caption{Impact of the hierarchical DNF encoder}
    \label{fig:dnf-curve}
  \end{subfigure}
  \hfill
  \begin{subfigure}{0.49\columnwidth}
    \centering
    \includegraphics[width=\textwidth]{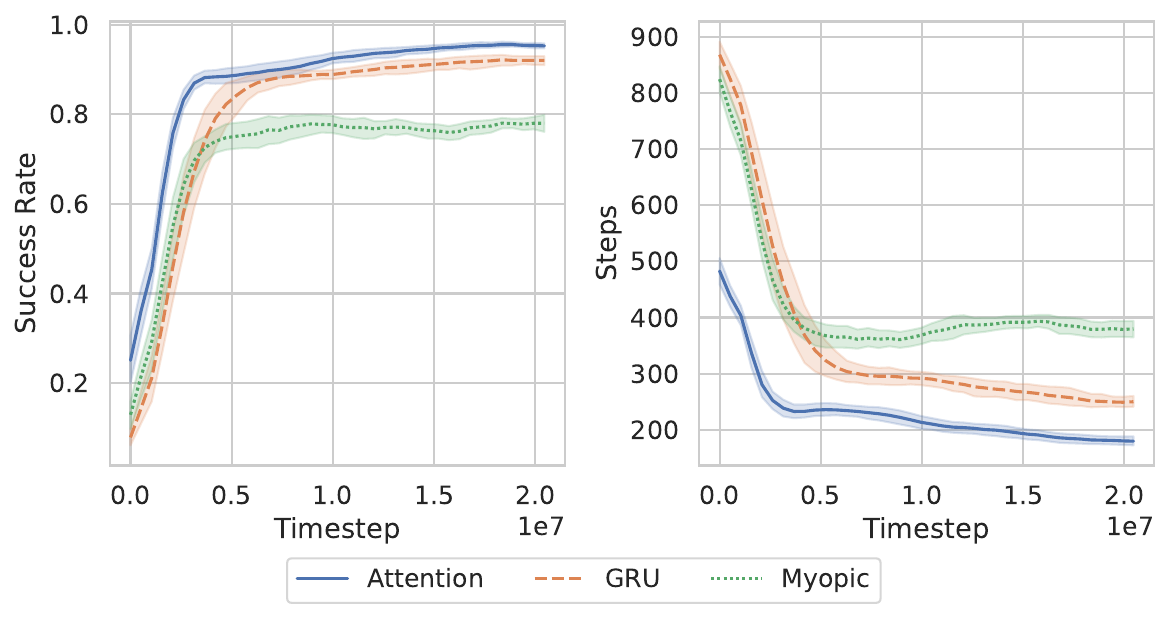}
    \caption{Impact of the temporal attention mechanism}
    \label{fig:att-curve}
  \end{subfigure}
  \begin{subfigure}{0.49\columnwidth}
    \centering
    \includegraphics[width=\textwidth]{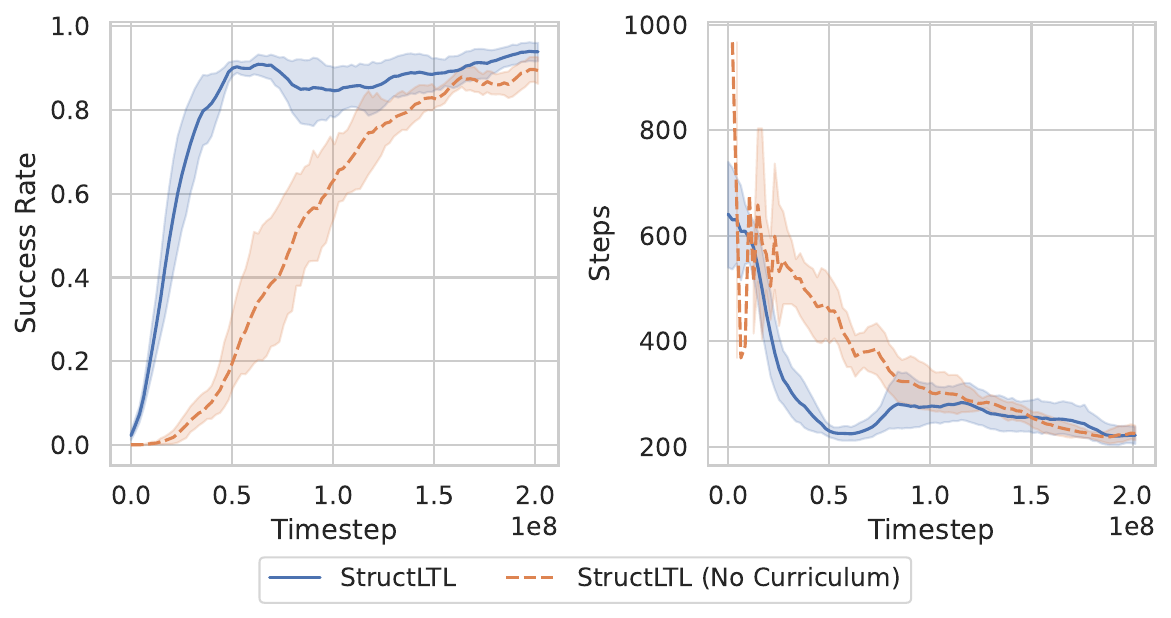}
    \caption{Impact of curriculum learning}
    \label{fig:curriculum-curve}
  \end{subfigure}
  \caption{Ablation studies. We report averages over finite-horizon evaluation tasks, computed over 50 episodes per task. Shaded areas indicate 95\% confidence intervals over 10 random seeds.}
  \label{fig:ablation-curves}
\end{figure*}

\subsubsection{Impact of Hierarchical DNF Encoder}
In this ablation study, we investigate the impact of our proposed hierarchical DNF encoder. In particular, we replace the DNF encoder with (i) a flat sequence model that processes Boolean formulae as lists of tokens and (ii) a generic GNN operating on syntax trees\@. For the sequence model, we use a gated recurrent unit (GRU)~\citep{cho2014Properties} with a token embedding size of 16, as in our main experiments. The GNN is instantiated as a graph convolutional network (GCN)~\citep{kipf2017Semisupervised} with embedding dimension 16, three layers (matching the maximum syntax tree height), and ReLU activations; the final representation of the formula is the final embedding of the root node. Note that we only replace the encoding of Boolean formulae themselves, in order to precisely ablate this part of our proposed architecture; the higher-level sequence encoding is still accomplished with our proposed attention mechanism.

The results in \cref{fig:dnf-curve} show the performance on Warehouse. They clearly demonstrate that our hierarchical DNF encoding scheme is superior to both flat token-based and GNN encodings of the Boolean formula. While token-based encodings show some success, the policy fails to learn useful representations as it progresses through the curriculum and encounters more complex Boolean formulae. The GNN initially learns quickly, but the representations fail to remain stable throughout training. In contrast, our hierarchical encoding mechanism achieves high success rates.

\subsubsection{Impact of Temporal Attention Mechanism}
We proceed to investigate the impact of the temporal attention mechanism. We compare our proposed approach to (i) a myopic baseline that only conditions on the first step of the Boolean formula sequence, and (ii) encoding the sequence of Boolean formulae with a GRU instead of attention. We conduct this ablation study in the ZoneEnv-NM environment.

\cref{fig:att-curve} shows that our proposed attention mechanism performs better than the alternatives. The myopic baseline initially learns quickly, but then plateaus. The GRU converges above the myopic baseline, but below the attention method.

\subsection{Impact of Curriculum Learning}\label{app:curriculum-learning}
Finally, we analyse the impact of curriculum learning. We compare training with our proposed curriculum on WarehouseEnv to training only on formula sequences sampled from the final curriculum stage. \cref{fig:curriculum-curve} shows that the curriculum significantly improves sample efficiency and leads to increased final performance. However, our proposed architecture is still able to learn a successful policy when not guided by a curriculum.

\section{Complexity Analysis}\label{app:complexity}
Let $n = |AP|$, $|\varphi|$ be the formula length, $T$ the sequence length, and $M$ the total number of literals in the formula sequence. Like other automata-based methods (e.g.\ DeepLTL, GenZ-LTL), we rely on constructing Büchi automata, which is doubly exponential in $|\varphi|$ in the worst case.
The complexity of the DFS to find all paths to accepting cycles depends on the number of simple paths from the current state (see \cite{abate2026Semantically} for a detailed analysis).
Constructing Boolean formulae using the Quine-McCluskey algorithm is exponential in $n$ in the worst case, but in practice restricted by the number of \textit{valid} assignments (e.g.\ the agent cannot be in both region\_a and region\_b at the same time). In terms of automaton size, formula construction is linear, since we need to construct one formula per transition in the worst case.

Crucially, all of this is precomputed once per test specification rather than at every environment step. Once the formulae have been constructed, the DNF encoder is linear in $M$ and the attention operation is linear in $T$ (not quadratic, since only the first step performs attention to the other steps).

\cref{tab:warehouse-runtime} shows the mean wall-clock time for each processing stage in WarehouseEnv across different sequence depths, averaged over 100 trials. The LDBA construction dominates the preprocessing time, and Boolean formula generation scales linearly with the number of transitions, as expected.

\begin{table}[t]
  \centering
  \caption{Mean wall-clock time (ms) over 100 trials for each processing stage on WarehouseEnv.}
  \label{tab:warehouse-runtime}
  \begin{tabular}{@{}lrrrr@{}}
    \toprule
    Stage                      & Depth 2 & Depth 4 & Depth 8 & Depth 12 \\
    \midrule
    LDBA construction          & 354.1   & 373.5   & 398.7   & 421      \\
    Boolean formula generation & 3.0     & 3.2     & 6.3     & 12.8     \\
    StructLTL forward pass     & 0.2     & 0.2     & 0.2     & 0.2      \\
    \bottomrule
  \end{tabular}
\end{table}

\section{Trajectory Visualisations}\label{app:trajectories}
\begin{figure*}[b!]
  \begin{subfigure}{\textwidth}
    \includegraphics[width=\textwidth]{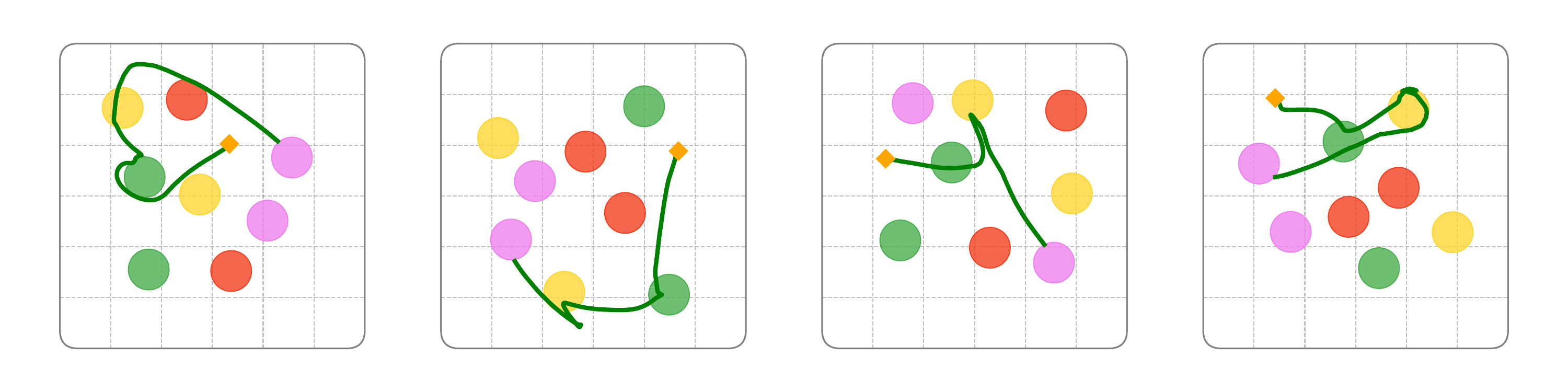}
    \caption{$\event (\mathsf{green} \wedge (\neg\mathsf{red} \until \mathsf{yellow})) \wedge \event \mathsf{purple}$}
  \end{subfigure}

  \bigskip

  \begin{subfigure}{\textwidth}
    \includegraphics[width=\textwidth]{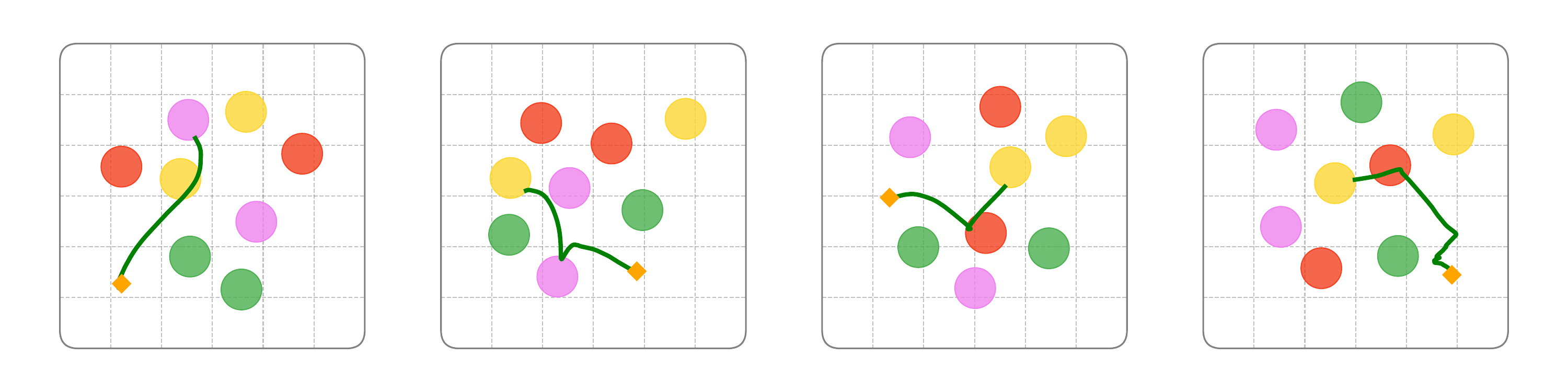}
    \caption{$\event (\mathsf{red} \lor \mathsf{purple}) \wedge \event \mathsf{yellow} \wedge \always\neg\mathsf{green}$}
  \end{subfigure}
  \bigskip

  \begin{subfigure}{\textwidth}
    \includegraphics[width=\textwidth]{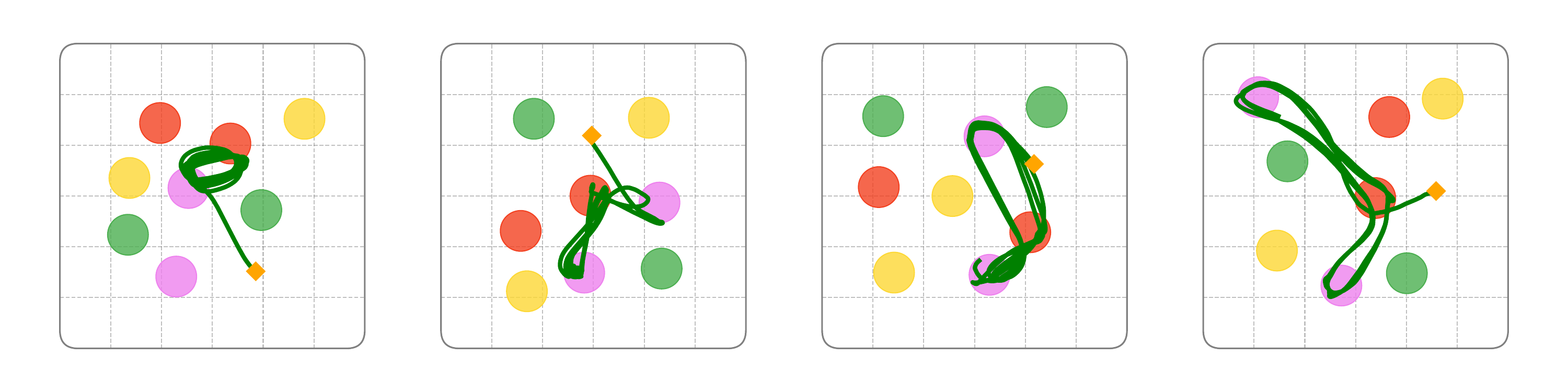}
    \caption{$\always\event\mathsf{purple}\land\always\event\mathsf{red}\land\always\neg\mathsf{yellow}$}
  \end{subfigure}
  \bigskip

  \begin{subfigure}{\textwidth}
    \includegraphics[width=\textwidth]{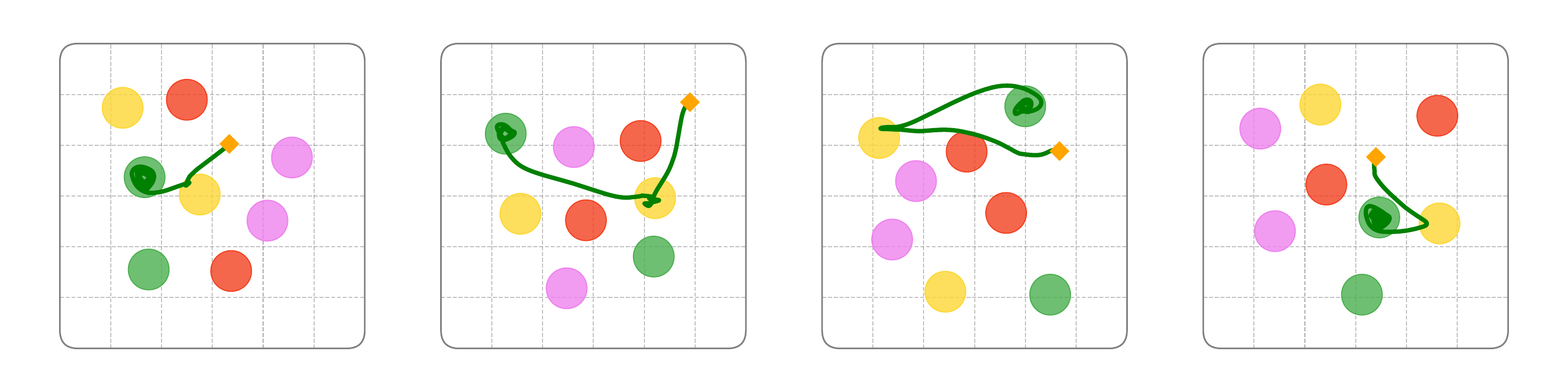}
    \caption{$\event\mathsf{yellow}\land\event\always\mathsf{green}\land\always(\mathsf{\mathsf{purple}\rightarrow\event\mathsf{red}})$}
  \end{subfigure}
  \caption{Example trajectories of StructLTL in ZoneEnv-NM. The orange diamond indicates the starting position of the agent. All trajectories were generated by the same policy after training.}
  \label{fig:zones-trajs}
\end{figure*}

\begin{figure*}[b!]
  \begin{subfigure}{\textwidth}
    \includegraphics[width=\textwidth]{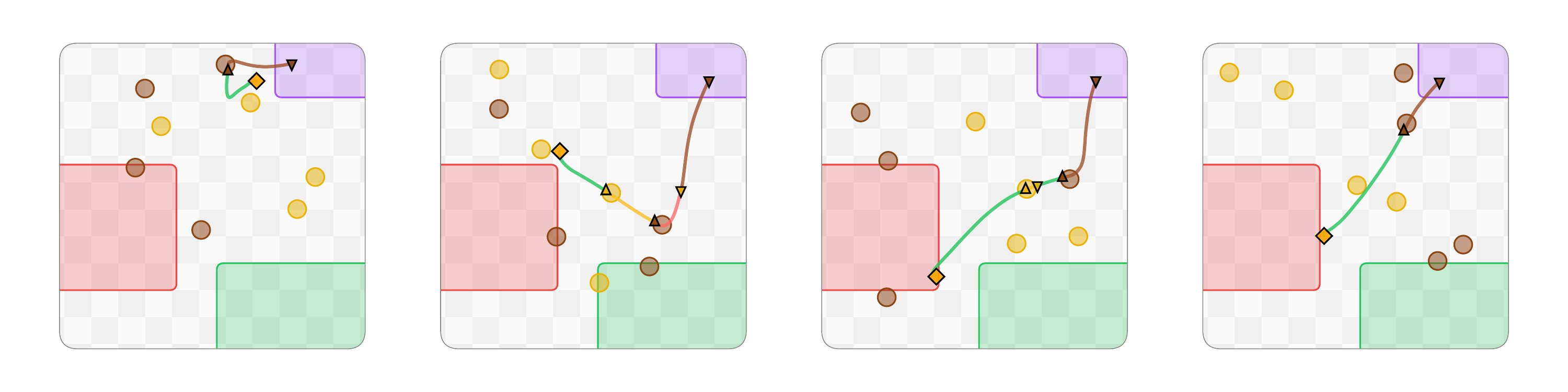}
    \caption{$\event (\mathsf{crate} \wedge \mathsf{door} \wedge (\mathsf{door} \until (\neg \mathsf{crate} \wedge \mathsf{door})))$}
  \end{subfigure}

  \bigskip

  \begin{subfigure}{\textwidth}
    \includegraphics[width=\textwidth]{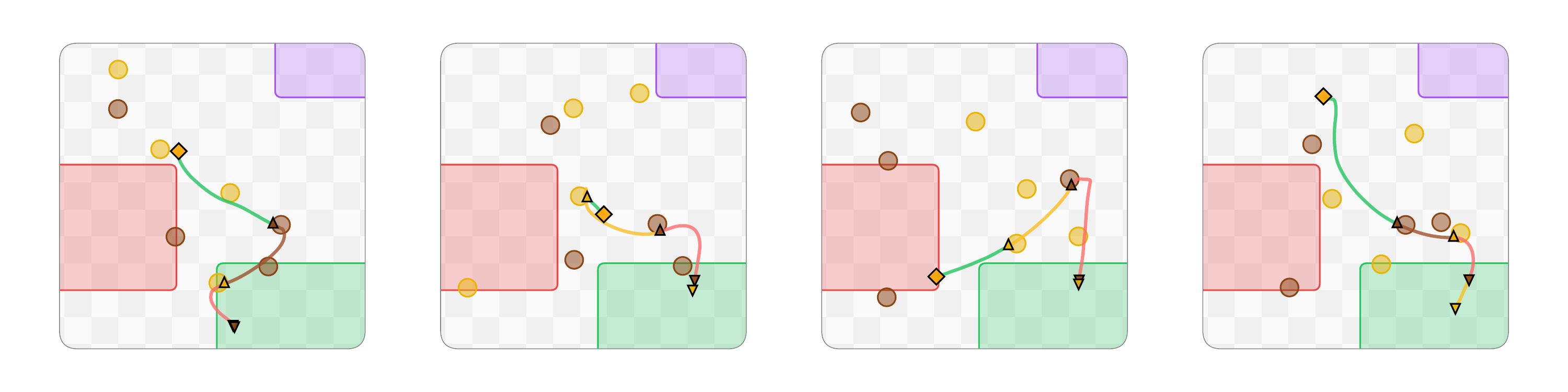}
    \caption{$\event (\mathsf{vase} \wedge \mathsf{crate} \wedge \mathsf{region\_b} \wedge (\mathsf{region\_b} \until (\neg \mathsf{vase} \wedge \neg \mathsf{crate} \wedge \mathsf{region\_b})))$}
  \end{subfigure}
  \bigskip

  \begin{subfigure}{\textwidth}
    \includegraphics[width=\textwidth]{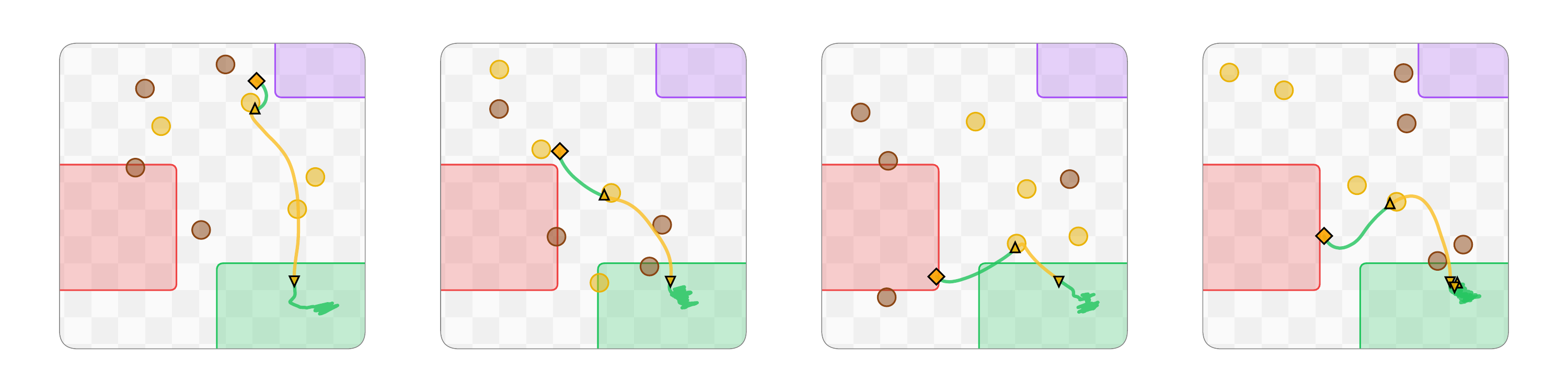}
    \caption{$\event (\mathsf{vase} \wedge \mathsf{region\_b} \wedge (\mathsf{region\_b} \until (\neg \mathsf{vase} \wedge \mathsf{region\_b}))) \wedge \event \always \mathsf{region\_b}$}
  \end{subfigure}
  \bigskip

  \begin{subfigure}{\textwidth}
    \includegraphics[width=\textwidth]{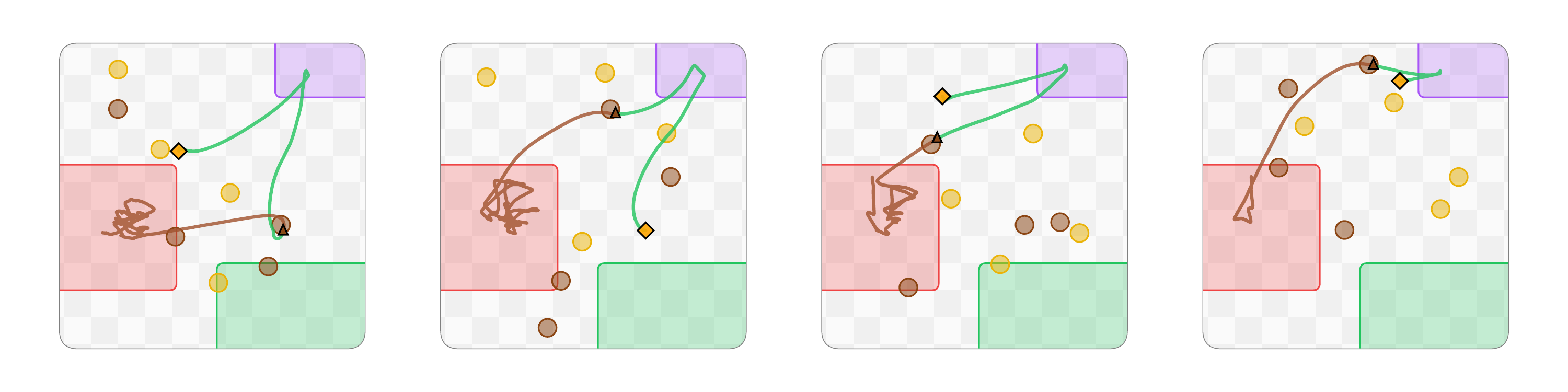}
    \caption{$\event\mathsf{door}\land \event\always(\mathsf{crate}\land\mathsf{region\_a})$}
  \end{subfigure}
  \caption{Example trajectories of StructLTL in the Warehouse environment. The orange diamond indicates the starting position of the agent, and yellow and brown circles correspond to vases and crates, respectively. A triangle pointing upwards indicates picking up an object of the corresponding colour, while a triangle pointing downwards indicates dropping an object. Different trajectory colours similarly highlight the objects the agent is currently carrying. All trajectories were generated by the same policy after training.}
  \label{fig:warehouse-trajs}
\end{figure*}

\cref{fig:zones-trajs,fig:warehouse-trajs} show trajectories produced by StructLTL on various specifications in ZoneEnv-NM and Warehouse, respectively.
The trajectories were obtained by executing the final policy at the end of training conditioned on the given tasks, and confirm that the policy correctly satisfies diverse specifications.

\begin{table}[ht]
\caption{Complete list of evaluation specifications.}
\centering
\small 
\label{tab:ltl_specs}
\resizebox{\textwidth}{!}{%
\begin{tabular}{l l p{12cm}}
\toprule
{ID} & {Environment} & {LTL Formula} \\
\midrule
\multicolumn{3}{c}{\textit{Finite Horizon}} \\
\midrule
$\varphi_{1}$ & ZoneEnv-NM & $(\event \mathsf{red}) \wedge (\neg \mathsf{red} \until (\mathsf{green} \wedge \event \mathsf{yellow}))$ \\
$\varphi_{2}$ & ZoneEnv-NM & $\event (\mathsf{red} \vee \mathsf{green}) \wedge \event \mathsf{yellow} \wedge \event \mathsf{purple}$ \\
$\varphi_{3}$ & ZoneEnv-NM & $\neg (\mathsf{purple} \vee \mathsf{yellow}) \until (\mathsf{red} \wedge \event \mathsf{green})$ \\
$\varphi_{4}$ & ZoneEnv-NM & $\neg \mathsf{green} \until ((\mathsf{red} \vee \mathsf{purple}) \wedge (\neg \mathsf{green} \until \mathsf{yellow}))$ \\
$\varphi_{5}$ & ZoneEnv-NM & $((\mathsf{green} \vee \mathsf{red}) \rightarrow (\neg \mathsf{yellow} \until \mathsf{purple})) \until \mathsf{yellow}$ \\
$\varphi_{6}$ & ZoneEnv-NM & $\event (\mathsf{green} \wedge (\neg \mathsf{red} \until \mathsf{yellow})) \wedge \event \mathsf{purple}$ \\
$\varphi_{7}$ & ZoneEnv-NM & $\event (\mathsf{yellow} \vee \mathsf{purple} \vee \mathsf{red}) \wedge (\neg \mathsf{yellow} \until \mathsf{green})$ \\
$\varphi_{8}$ & ZoneEnv-NM & $\event \mathsf{purple} \wedge \event \mathsf{green} \wedge \event \mathsf{yellow} \wedge \event \mathsf{red}$ \\
$\varphi_{9}$ & Warehouse & $\event (\mathsf{region\_a} \wedge \event (\mathsf{crate} \wedge \event \mathsf{region\_b}))$ \\
$\varphi_{10}$ & Warehouse & $\neg (\mathsf{region\_a} \vee \mathsf{door}) \until (\mathsf{vase} \wedge \mathsf{region\_b})$ \\
$\varphi_{11}$ & Warehouse & $\neg \mathsf{door} \until (\mathsf{region\_a} \wedge (\neg \mathsf{vase} \until \mathsf{region\_b}))$ \\
$\varphi_{12}$ & Warehouse & $\neg (\mathsf{vase} \vee \mathsf{crate}) \until (\mathsf{region\_a} \wedge \event (\mathsf{door} \wedge \event \mathsf{region\_b}))$ \\
$\varphi_{13}$ & Warehouse & $\event (\mathsf{vase} \wedge \mathsf{region\_a} \wedge (\mathsf{region\_a} \until (\neg \mathsf{vase} \wedge \mathsf{region\_a})))$ \\
$\varphi_{14}$ & Warehouse & $\event (\mathsf{vase} \wedge \mathsf{crate} \wedge \mathsf{region\_b} \wedge (\mathsf{region\_b} \until (\neg \mathsf{vase} \wedge \neg \mathsf{crate} \wedge \mathsf{region\_b})))$ \\
$\varphi_{15}$ & Warehouse & $(\mathsf{crate} \rightarrow (\mathsf{crate} \until \mathsf{region\_b})) \until (\mathsf{crate} \wedge \mathsf{region\_b} \wedge (\mathsf{region\_b} \until (\neg \mathsf{crate} \wedge \mathsf{region\_b})))$ \\
$\varphi_{16}$ & Warehouse & $(\mathsf{vase} \rightarrow (\mathsf{vase} \until \mathsf{door})) \until (\event (\mathsf{crate} \wedge \mathsf{region\_b} \wedge (\mathsf{region\_b} \until (\neg \mathsf{crate} \wedge \mathsf{region\_b}))) \wedge \event (\mathsf{vase} \wedge \mathsf{door} \wedge (\mathsf{door} \until (\neg \mathsf{vase} \wedge \mathsf{door}))))$ \\
$\varphi_{17}$ & Warehouse & $(\mathsf{crate} \rightarrow (\neg (\mathsf{region\_a} \vee \mathsf{region\_b})) \until \mathsf{door}) \until (\mathsf{vase} \wedge \mathsf{door} \wedge (\mathsf{door} \until (\neg \mathsf{vase} \wedge \mathsf{door})))$ \\
$\varphi_{18}$ & Warehouse & $\event (\mathsf{crate} \wedge \mathsf{region\_a} \wedge (\mathsf{region\_a} \until (\neg \mathsf{crate} \wedge \mathsf{region\_a}))) \wedge \event (\mathsf{vase} \wedge \mathsf{door} \wedge (\mathsf{door} \until (\neg \mathsf{vase} \wedge \mathsf{door})))$ \\
\midrule
\multicolumn{3}{c}{\textit{Infinite Horizon}} \\
\midrule
$\psi_{1}$ & ZoneEnv-NM & $\always \event \mathsf{red} \wedge \always \event \mathsf{green} \wedge \always \event \mathsf{yellow}$ \\
$\psi_{2}$ & ZoneEnv-NM & $\event ((\mathsf{red} \vee \mathsf{purple}) \wedge \event \mathsf{yellow}) \wedge \always \event \mathsf{green} \wedge \always \event \mathsf{red}$ \\
$\psi_{3}$ & ZoneEnv-NM & $\event (\mathsf{red} \wedge \event (\mathsf{green} \vee \mathsf{purple})) \wedge \always (\mathsf{red} \rightarrow \event (\mathsf{green} \wedge \event \mathsf{yellow})) \wedge \always (\mathsf{yellow} \rightarrow \event \mathsf{red})$ \\
$\psi_{4}$ & ZoneEnv-NM & $\event \always \mathsf{red}$ \\
$\psi_{5}$ & ZoneEnv-NM & $\event \always \mathsf{red} \wedge \event (\mathsf{yellow} \wedge \event \mathsf{green})$ \\
$\psi_{6}$ & ZoneEnv-NM & $\event \always \mathsf{purple} \wedge \always \neg \mathsf{yellow}$ \\
$\psi_{7}$ & ZoneEnv-NM & $\always ((\mathsf{green} \vee \mathsf{yellow}) \rightarrow \event \mathsf{red}) \wedge \event \always (\mathsf{green} \vee \mathsf{purple})$ \\
$\psi_{8}$ & ZoneEnv-NM & $\event (\mathsf{purple} \wedge \event (\mathsf{green} \vee \mathsf{red})) \wedge \event \always \mathsf{yellow}$ \\
$\psi_{9}$ & Warehouse & $\always \event \mathsf{region\_a} \wedge \always \event \mathsf{region\_b} \wedge \always \event \mathsf{door}$ \\
$\psi_{10}$ & Warehouse & $\always \event (\mathsf{vase} \wedge \event \neg \mathsf{vase}) \wedge \always \event (\mathsf{crate} \wedge \event \neg \mathsf{crate}) \wedge \always \neg \mathsf{region\_a}$ \\
$\psi_{11}$ & Warehouse & $\event (\mathsf{vase} \wedge \mathsf{region\_a} \wedge (\mathsf{region\_a} \until (\neg \mathsf{vase} \wedge \mathsf{region\_a}))) \wedge \always \event \mathsf{door} \wedge \always \event \mathsf{region\_b}$ \\
$\psi_{12}$ & Warehouse & $\always \event (\mathsf{vase} \wedge \mathsf{region\_a} \wedge (\mathsf{region\_a} \until (\neg \mathsf{vase} \wedge \mathsf{region\_a}))) \wedge \always \event (\mathsf{crate} \wedge \mathsf{region\_b} \wedge (\mathsf{region\_b} \until (\neg \mathsf{crate} \wedge \mathsf{region\_b})))$ \\
$\psi_{13}$ & Warehouse & $\event \always \mathsf{region\_a}$ \\
$\psi_{14}$ & Warehouse & $\event (\mathsf{vase} \wedge \mathsf{region\_b} \wedge (\mathsf{region\_b} \until (\neg \mathsf{vase} \wedge \mathsf{region\_b}))) \wedge \event \always \mathsf{region\_b}$ \\
$\psi_{15}$ & Warehouse & $\event \always (\mathsf{crate} \wedge \mathsf{region\_a})$ \\
$\psi_{16}$ & Warehouse & $\event \always (\mathsf{vase} \wedge \mathsf{region\_b}) \wedge \always \neg \mathsf{region\_a} \wedge \always \neg \mathsf{crate}$ \\
$\psi_{17}$ & Warehouse & $\always ((\mathsf{door} \vee \mathsf{region\_b}) \rightarrow \event \mathsf{region\_a}) \wedge \event \always (\mathsf{vase} \vee \mathsf{crate})$ \\
$\psi_{18}$ & Warehouse & $\event \always \mathsf{crate} \wedge \always (\mathsf{vase} \rightarrow \event (\mathsf{door} \wedge \neg \mathsf{crate}))$ \\
\bottomrule
\end{tabular}
}
\end{table}
\begin{table}[h]
  \centering
  \caption{Evaluation specifications used in the compositional generalisation experiment.}
  \begin{tabular}{ll}
    \toprule
    {ID}           & {LTL Formula}                                                                                                                                                                  \\
    \midrule
    $\varphi_{19}$ & $\event (\mathsf{crate} \land \mathsf{vase} \land \mathsf{region\_a})$                                                                                                         \\
    $\varphi_{20}$ & $\event (\mathsf{crate} \land \mathsf{vase} \land \mathsf{region\_b})$                                                                                                         \\
    $\varphi_{21}$ & $\event (\mathsf{crate} \land \mathsf{vase} \land \mathsf{door})$                                                                                                              \\
    $\varphi_{22}$ & $\event (\mathsf{vase} \land \mathsf{crate} \land \mathsf{region\_a} \land (\mathsf{region\_a} \until (\neg\mathsf{vase} \land \neg\mathsf{crate} \land \mathsf{region\_a})))$ \\
    $\varphi_{23}$ & $\event (\mathsf{vase} \land \mathsf{crate} \land \mathsf{region\_b} \land (\mathsf{region\_b} \until (\neg\mathsf{vase} \land \neg\mathsf{crate} \land \mathsf{region\_b})))$ \\
    $\varphi_{24}$ & $\event (\mathsf{vase} \land \mathsf{crate} \land \mathsf{door} \land (\mathsf{door} \until (\neg\mathsf{vase} \land \neg\mathsf{crate} \land \mathsf{door})))$                \\
    \bottomrule
  \end{tabular}
  \label{tab:gen_formulae}
\end{table}

\clearpage

\clearpage
\section*{NeurIPS Paper Checklist}

\begin{enumerate}

    \item {\bf Claims}
    \item[] Question: Do the main claims made in the abstract and introduction accurately reflect the paper's contributions and scope?
    \item[] Answer: \answerYes{} 
    \item[] Justification: The abstract and introduction accurately reflect the paper's contributions and scope, including an explicit list of contributions in the introduction. All claims are supported by the experiments presented in the paper.

    \item {\bf Limitations}
    \item[] Question: Does the paper discuss the limitations of the work performed by the authors?
    \item[] Answer: \answerYes{} 
    \item[] Justification: We include a discussion of limitations in \cref{sec:conclusion_limitations}.

    \item {\bf Theory assumptions and proofs}
    \item[] Question: For each theoretical result, does the paper provide the full set of assumptions and a complete (and correct) proof?
    \item[] Answer: \answerNA{} 
    \item[] Justification: The paper does not include theoretical results.

    \item {\bf Experimental result reproducibility}
    \item[] Question: Does the paper fully disclose all the information needed to reproduce the main experimental results of the paper to the extent that it affects the main claims and/or conclusions of the paper (regardless of whether the code and data are provided or not)?
    \item[] Answer: \answerYes{} 
    \item[] Justification: The paper includes comprehensive experimental details in Appendix \ref{app:experimental_details}, and all code used for experiments can be found in the supplementary material.

    \item {\bf Open access to data and code}
    \item[] Question: Does the paper provide open access to the data and code, with sufficient instructions to faithfully reproduce the main experimental results, as described in supplemental material?
    \item[] Answer: \answerYes{} 
    \item[] Justification: All code used for experiments can be found in the supplementary material.

    \item {\bf Experimental setting/details}
    \item[] Question: Does the paper specify all the training and test details (e.g., data splits, hyperparameters, how they were chosen, type of optimizer) necessary to understand the results?
    \item[] Answer: \answerYes{} 
    \item[] Justification: The paper includes comprehensive experimental details in \cref{app:experimental_details}. These details can also be found in the code provided in the supplementary material.

    \item {\bf Experiment statistical significance}
    \item[] Question: Does the paper report error bars suitably and correctly defined or other appropriate information about the statistical significance of the experiments?
    \item[] Answer: \answerYes{} 
    \item[] Justification: Details on error bars for figures and tables are provided in \cref{sec:experiments}, under ``Evaluation Protocol''. Further technical details can be found in the code provided in the supplementary material.

    \item {\bf Experiments compute resources}
    \item[] Question: For each experiment, does the paper provide sufficient information on the computer resources (type of compute workers, memory, time of execution) needed to reproduce the experiments?
    \item[] Answer: \answerYes{} 
    \item[] Justification: Details on compute resources are provided in \cref{app:implementation}.

    \item {\bf Code of ethics}
    \item[] Question: Does the research conducted in the paper conform, in every respect, with the NeurIPS Code of Ethics \url{https://neurips.cc/public/EthicsGuidelines}?
    \item[] Answer: \answerYes{} 
    \item[] Justification: The authors have read and followed the NeurIPS Code of Ethics. This work does not involve human subjects, sensitive data, or applications with foreseeable harmful misuse.

    \item {\bf Broader impacts}
    \item[] Question: Does the paper discuss both potential positive societal impacts and negative societal impacts of the work performed?
    \item[] Answer: \answerNA{} 
    \item[] Justification: This paper presents foundational research on Machine Learning. It is not tied to any specific application or deployment, and there is no direct path to negative societal impacts.

    \item {\bf Safeguards}
    \item[] Question: Does the paper describe safeguards that have been put in place for responsible release of data or models that have a high risk for misuse (e.g., pre-trained language models, image generators, or scraped datasets)?
    \item[] Answer: \answerNA{} 
    \item[] Justification: This paper poses no such risks. It does not release pre-trained models, image generators, or scraped datasets.

    \item {\bf Licenses for existing assets}
    \item[] Question: Are the creators or original owners of assets (e.g., code, data, models), used in the paper, properly credited and are the license and terms of use explicitly mentioned and properly respected?
    \item[] Answer: \answerYes{} 
    \item[] Justification: This paper uses JAX (Apache v2.0) and Rabinizer 4 (GNU GPL v3.0), both cited in \cref{app:implementation}. Neither is bundled with our code; JAX is installed as a Python dependency and Rabinizer is installed separately by the user.

    \item {\bf New assets}
    \item[] Question: Are new assets introduced in the paper well documented and is the documentation provided alongside the assets?
    \item[] Answer: \answerYes{} 
    \item[] Justification: All code used for experiments is publicly released at \href{https://github.com/mathiasj33/structltl}{github.com/mathiasj33/structltl}. It includes documentation and instructions for reproducing all experiments.

    \item {\bf Crowdsourcing and research with human subjects}
    \item[] Question: For crowdsourcing experiments and research with human subjects, does the paper include the full text of instructions given to participants and screenshots, if applicable, as well as details about compensation (if any)?
    \item[] Answer: \answerNA{} 
    \item[] Justification: This paper does not involve crowdsourcing nor research with human subjects.

    \item {\bf Institutional review board (IRB) approvals or equivalent for research with human subjects}
    \item[] Question: Does the paper describe potential risks incurred by study participants, whether such risks were disclosed to the subjects, and whether Institutional Review Board (IRB) approvals (or an equivalent approval/review based on the requirements of your country or institution) were obtained?
    \item[] Answer: \answerNA{} 
    \item[] Justification: This paper does not involve crowdsourcing nor research with human subjects.

    \item {\bf Declaration of LLM usage}
    \item[] Question: Does the paper describe the usage of LLMs if it is an important, original, or non-standard component of the core methods in this research? Note that if the LLM is used only for writing, editing, or formatting purposes and does \emph{not} impact the core methodology, scientific rigor, or originality of the research, declaration is not required.
    \item[] Answer: \answerNA{} 
    \item[] Justification: The core method development in this paper does not involve LLMs as any important, original, or non-standard components.

\end{enumerate}

\end{document}